\documentclass[lettersize,journal]{IEEEtran}
\usepackage{amsmath,amsfonts}
\usepackage{algorithmic}
\usepackage{algorithm}
\usepackage{array}
\usepackage[caption=false,font=normalsize,labelfont=sf,textfont=sf]{subfig}
\usepackage{textcomp}
\usepackage{stfloats}
\usepackage{url}
\usepackage{verbatim}
\usepackage{graphicx}
\usepackage{wrapfig}
\usepackage{cite}
\usepackage{soul} 
\usepackage{color, xcolor} 
\usepackage{url}
\usepackage{tabularx}
\usepackage{multicol}
\usepackage{multirow}
\usepackage{tabularray}
\usepackage{makecell}
\usepackage{booktabs}
\usepackage{subcaption}
\usepackage{xcolor}

\usepackage{orcidlink}
\usepackage[misc]{ifsym} 
\begin{document}

\title{Efficient Orchestrated AI Workflows Execution on Scale-out Spatial Architecture}

\author{
	\IEEEauthorblockN{
		Jinyi Deng\orcidlink{0000-0001-8666-8463},
        ~\IEEEmembership{Student Member,~IEEE},
        Xinru Tang\orcidlink{0009-0004-6038-3709},
        Zhiheng Yue\orcidlink{0000-0003-4084-3478},
        Guangyang Lu\orcidlink{0009-0008-3523-1649},
        Qize Yang\orcidlink{0009-0006-2221-4706},
        Jiahao Zhang\orcidlink{0009-0000-8379-7489},
        Jinxi Li\orcidlink{0009-0002-4841-3798},
        Chao Li\orcidlink{0000-0001-6218-4659},
        ~\IEEEmembership{Senior Member,~IEEE},
        Shaojun Wei\orcidlink{0000-0001-5117-7920},
        ~\IEEEmembership{Fellow,~IEEE},
        Yang Hu\orcidlink{0000-0001-6942-4395},
        ~\IEEEmembership{Member,~IEEE},
        and Shouyi Yin\orcidlink{0000-0003-2309-572X}
        ~\IEEEmembership{Senior Member,~IEEE}\\
    }
    \thanks{

This work was supported in part by the NSFC under Grant 62125403, and Grant 92164301; in part by the National Science and Technology Major Project under Grant 2022ZD0115201; in part by the National Key Research and Development Program under Grant 2021ZD0114400; in part by the Beijing S\&T Project Z221100007722023; in part by the Beijing National Research Center for Information Science and Technology; and in part by the Beijing Advanced Innovation Center for Integrated Circuits. 

Jinyi Deng, Xinru Tang, Zhiheng Yue, Guangyang Lu, Qize Yang, Jiahao Zhang, Jinxi Li, Shaojun Wei, Yang Hu and Shouyi Yin are with School of Integrated Circuits, Tsinghua University, Beijing 100084, China. E-mail:\{dengjy18, tangxr23, yuezh20, lugy23, yqz23, jhzhang23, ljx23\}@mails.tsinghua.edu.cn, \{wsj, hu\_yang, yinsy\}@tsinghua.edu.cn.

Chao Li are with the Shanghai Jiao Tong University, Shanghai 200240, China. E-mail: lichao@cs.sjtu.edu.cn. 

Manuscript received April 30, 2024.
}
}



\maketitle

\begin{abstract}



Given the increasing complexity of AI applications, traditional spatial architectures frequently fall short. Our analysis identifies a pattern of interconnected, multi-faceted tasks encompassing both AI and general computational processes. 
In response, we have conceptualized "Orchestrated AI Workflows," an approach that integrates various tasks with logic-driven decisions into dynamic, sophisticated workflows.
Specifically, we find that the intrinsic Dual Dynamicity of Orchestrated AI Workflows, namely dynamic execution times and frequencies of Task Blocks, can be effectively represented using the Orchestrated Workflow Graph.
Furthermore, the intrinsic Dual Dynamicity poses challenges to existing spatial architecture, namely Indiscriminate Resource Allocation, Reactive Load Rebalancing, and Contagious PEA Idleness. 

To overcome these challenges, we present Octopus, a scale-out spatial architecture and a suite of advanced scheduling strategies optimized for executing Orchestrated AI Workflows, such as the Discriminate Dual-Scheduling Mechanism, Adaptive TBU Scheduling Strategy, and Proactive Cluster Scheduling Strategy. Our evaluations demonstrate that Octopus significantly outperforms traditional architectures in handling the dynamic demands of Orchestrated AI Workflows, and possesses robust scalability in large scale hardware such as wafer-scale chip.

\end{abstract}

\begin{IEEEkeywords}
Orchestrated AI Workflows, Dual Dynamicity, Spatial Architecture, Dynamic Schedule, Wafer-Scale Scalability.
\end{IEEEkeywords}

\section{Introduction}\label{sec:intro}

\begin{figure}[t]
    \setlength{\abovecaptionskip}{0pt}
    \centering
    \includegraphics[width=\linewidth]{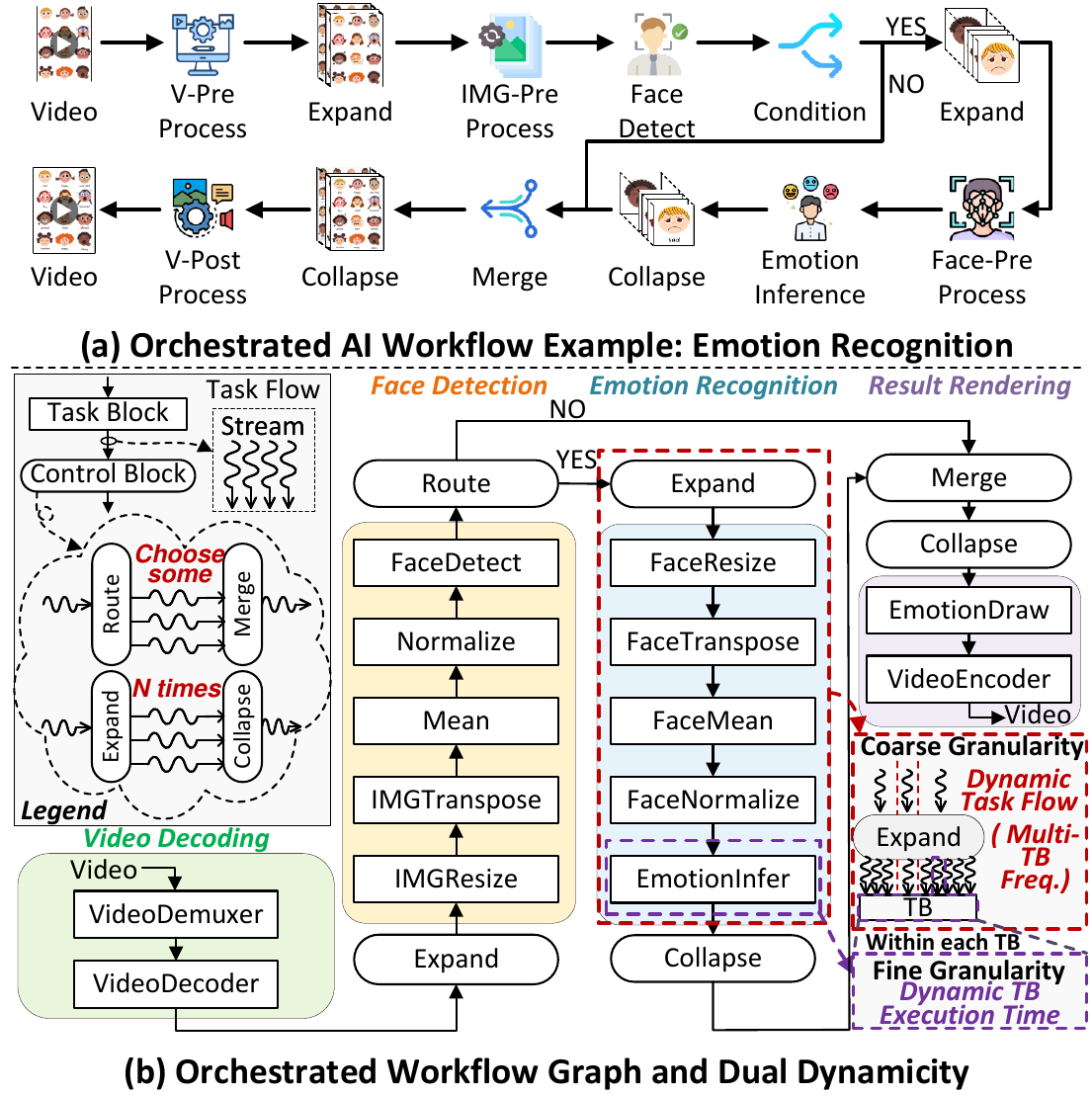}
    \caption{Orchestrated AI Workflow Example: Emotion Recognition; Orchestrated Workflow Graph Extracting the Inherent Logic and intrinsic characteristics of Orchestrated AI Workflow.}
    \label{fig1}
    \vspace{-20pt}
\end{figure} 

\IEEEPARstart{A}{s} demand for AI applications continues to become more complicated and specific \cite{gpt-4,cvcuda,distributed_zb8}, and the need for deeper-level processing and decision making increases, traditional processes based on a single AI model struggle to meet these requirements. 
AI models, along with general processing tasks, interconnect intricately through logical decisions, forming a powerful, yet sophisticated workflow. We name these emerging workloads as \textit{\textbf{Orchestrated AI Workflows}}. 
An example of an Orchestrated AI Workflow is emotion recognition, as illustrated in Figure \ref{fig1}. The output of a Face Detection model needs to undergo logical judgment, combined with other general processing, before being processed by Emotion Inference model.

Orchestrated AI Workflows are increasingly crucial across a wide array of commercial applications and sophisticated AI systems, playing a central role in functions ranging from Driver and Passenger Status Recognition to Optical Character Recognition \cite{ModelArts}.
Their importance is even more pronounced in the development of large multimodal models (LMM) such as Gemini \cite{geminiteam2024gemini} and Sora \cite{Sora}, which require intricate pre-processing of diverse media types for better decision making.
In recommendation systems utilized by Netflex \cite{Netflix} and Twitter \cite{Twitter}, Orchestrated AI Workflows play a crucial role in assimilating and prioritizing multiple AI models for processing by sophisticated ranking AI.
Furthermore, the Mixture of Experts (MoE) model \cite{MOE} epitomizes the Orchestrated AI Workflow by integrating multiple expert AI models with GateNet, which facilitates logical decision-making processes.

In our research, we are the first to present the \textit{Orchestrated Workflow Graph (OWG)} that systematically describes the inherent logical connections and embedded characteristics of Orchestrated AI Workflows. We employ the concept of Task Blocks (TB) to encapsulate AI and general computational tasks, while Control Blocks (CB) are used to delineate logical decision-making processes. Moreover, the data flow between these components is described using the Task Flow. Using OWG, we discover that Orchestrated AI Workflows exhibit what we call \textit{Dual Dynamicity}, which encompasses two significant aspects: {\textit{1) certain TBs exhibit dynamic execution times at a fine granularity}, and \textit{2) multiple TBs display dynamic execution frequencies at a coarse granularity}. 
Orchestrated AI Workflows stands out from traditional AI by exhibiting "Dual Dynamicity", which necessitates dynamically adjusting the allocation of hardware resources for different tasks in response to changing input data, unlike traditional AI where input variability does not alter the model's structure on the hardware. This attribute significantly influences the hardware requirements of Orchestrated AI Workflows.

Dual Dynamicity presents significant challenges to existing heterogeneous architectures\cite{jiang2024htdcr_zb9} during the execution of Orchestrated AI Workflows. A key limitation arises from the static nature of the compilation process. Once compiled, TBs cannot dynamically adjust to changing hardware resource demands during execution, which can lead to resource idleness.
Spatial architectures \cite{prabhakar2017plasticine,koeplinger2018spatial,swamy2023homunculus,ko2022accelerating,weng2020hybrid,zhang2019scalable,chen2016eyeriss,sankaralingam2022mozart,tan2022drips,karunaratne2017hycube,chin2017cgra,vasilyev2016evaluating,kou2022geml,kong2023mapzero,gobieski2021snafu,luo2023ml,adhi2022exploration,yoo2023lighttrader,dataflow_engine_zb1,reconfigurable_zb2,deng2022energy_zb3,wu2024swg_zb4,mapping_zb6}
are promising for addressing these challenges due to their flexible computational features, including reconfigurable processing element arrays (PEAs) for AI and computing tasks, enhanced by strong parallel processing capabilities. Additionally, spatial architectures expose low-level computational and network resources to hardware/software interfaces, enabling the potential for dynamic adjustments in resource allocation during runtime. Those address the Dual Dynamic requirements of Orchestrated AI workflows.

However, our observations indicate that existing spatial architectures fundamentally fail to perform efficient dynamic scheduling when managing the aforementioned Dual Dynamicity. This limitation stems from the fact that: 
(1) \textit{Indiscriminate Resource Allocation} arises because the current spatial architecture allocates schedulable tasks to two distinct exploitable PEA resources without differentiation;
(2) \textit{Reactive Load Rebalancing} involves awaiting threshold-triggered workload imbalances before responding, resulting in prolonged data transfers and heightened idleness;
and (3) \textit{Contagious PEA Idleness} refers to idle states spreading among interconnected PEAs due to a lack of timely rescheduling.
What makes it worse is that Orchestrated AI workflows tend more often to deploy on large-scale spatial architecture, such as chiplet system or even wafer-scale system \cite{oursurvey,dojo_wafer,UCLA_Wafer,Cerebras_Wafer,Cerebras_WSE}, the above mentioned challenges will exacerbate. 

To address the challenges outlined, we introduce Octopus, a novel scale-out spatial architecture and suite of scheduling strategies optimized for Orchestrated AI Workflow processing. Octopus encompasses Task Block Processing Units (TBUs) and Control Block Processing Units (CBUs). TBUs improve from basic PEAs to speed up Task Block computation. Concurrently, CBUs with distributed memory and routers have been optimized for more efficient Control Block processing. Octopus features scheduling strategies such as the \textit{Discriminate Dual-Scheduling Mechanism}, \textit{Adaptive TBU Scheduling Strategy}, and \textit{Proactive Cluster Scheduling Strategy}. 


The \textit{Discriminate Dual-Scheduling Mechanism} addresses the challenge of Indiscriminate Resource Allocation by introducing two separate atomic scheduling levels, enabling more precise coordination of tasks and resources at each dynamic layer. By segmenting OWGs into sub-OWGs at the CB, it identifies two levels: sub-OWG and TB for tasks, and Cluster and TBU for resources. Variations in sub-OWG execution frequencies are handled by coarse-grained Proactive Cluster Scheduling, while dynamic TB execution times are addressed with fine-grained Adaptive TBU Scheduling within clusters. 



The \textit{Adaptive TBU Scheduling Strategy} dynamically adjusts to TB execution times by continuously monitoring input and output streams, ensuring optimal task configuration. This approach leverages decentralized control semantics between CBUs and TBUs through the control flow plane, enabling agile transmission of control information, thus addressing the Contagious PEA Idleness challenge. 


The \textit{Proactive Cluster Scheduling Strategy}, synchronized with the Diffusive Load-Balancing Model, proactively addresses Reactive Load Rebalancing by adjusting task distributions in real-time. This strategy employs decentralized control and periodic inter-cluster communications, enabling clusters to autonomously redistribute tasks upon detecting imbalances and align load conditions for optimal equilibrium. This decentralized method reduces long-distance data transfers and network congestion, boosting system efficiency and responsiveness.


The contributions of this work are as follows:

\begin{enumerate}

\item \textbf{Orchestrated AI Workflow Presentation and Dual Dynamicity Identification:} We present Orchestrated AI Workflows, formalizing them through the Orchestrated Workflow Graph and identifying Dual Dynamicity to address changing execution demands dynamically.

\item \textbf{Challenges due to Dual Dynamicity in existing spatial architectures:} We identified challenges including Indiscriminate Resource Allocation, Reactive Load Rebalancing, and Contagious PEA Idleness which collectively hinder efficiency.

\item \textbf{Proposed Octopus Architecture and Scheduling Strategies:} Our novel scale-out spatial architecture features segmented processing units — TBUs and CBUs, supported by three adaptive scheduling strategies, optimizing task alignment and resource utilization.

\item \textbf{Comprehensive Implementation and Evaluation: }We implement a complete system that includes the compiler, simulator, and RTL design, rigorously testing Octopus against state-of-the-art spatial architectures. Results show Octopus outperforming by 2.50× on average and up to 4.26× across various Orchestrated AI Workflows. 
A case study further confirmed the scalability and effectiveness of our execution paradigm in a wafer-scale spatial architecture.
\end{enumerate}

\section{Workload Characterization}
In this section, to uncover the root cause of the challenges in implementing Orchestrated AI Workflows in spatial architectures, we provide a definition of the Orchestrated Workflow Graph (OWG) that fully articulates its underlying logical correlations and inherent features and summarize its Dual Dynamicity.


\subsection{Orchestrated AI Workflow}

An Orchestrated AI Workflow represents a comprehensive framework for AI applications that goes beyond basic single AI model scenarios.
It involves the systematic integration of \textit{multiple AI models, general processing tasks, and logical decisions that are intricately interwoven}.

Orchestrated AI Workflows, exemplified by Emotion Recognition, are critical and widespread. As shown in Figure \ref{fig1}, the Emotion Recognition process is segmented into four stages: video decoding, face detection, emotion recognition, and result rendering. This setup incorporates several AI models such as face detection and expression recognition, sequentially integrated for thorough video frame analysis. It encompasses general processing tasks like video decoding and rendering, which prepare data for further examination. Key decision points within the workflow optimize strategies for managing detected faces, improving efficiency and precision of the outcomes.

Orchestrated AI Workflows are increasingly vital across diverse commercial applications and advanced AI systems, crucial for efficiently processing. Everyday examples include Driver and Passenger Status Recognition, Street Flow Recognition, Crowd Mask Recognition, One-Shot Video Object Segmentation, and Optical Character Recognition \cite{ModelArts}, showcasing their essential role across diverse scenarios.


The significance of Orchestrated AI Workflows is magnified with the advent of Large Multimodal Models (LMMs), pivotal in advanced industrial AI applications. Models such as Gemini \cite{geminiteam2024gemini} and Sora \cite{Sora} demonstrate the necessity for complex preprocessing tasks across different media types like video and audio. These workflows integrate varied information, enhancing the coherence and efficiency of decision-making processes.

In the realm of recommendations systems, Orchestrated AI Workflows are pivotal. Netflex and Twitter Recommendation systems \cite{Netflix,Twitter} intricately blend models and select the most relevant information before it reaches a sophisticated heavy ranking AI, showcasing the orchestration in handling and prioritizing vast amounts of data.

Furthermore, Mixture of Experts (MoE) \cite{MOE} epitomizes the Orchestrated AI Workflow paradigm, combining multiple AI models with logical decision-making processes. It integrates multiple specialized sub-models as experts to handle different tasks, improving complex linguistic processing. Selectively activating relevant experts based on the input improves computational efficiency and reduces energy consumption.

\begin{figure}[t]

    \setlength{\abovecaptionskip}{0pt}
    \centering
    \includegraphics[width=\linewidth]{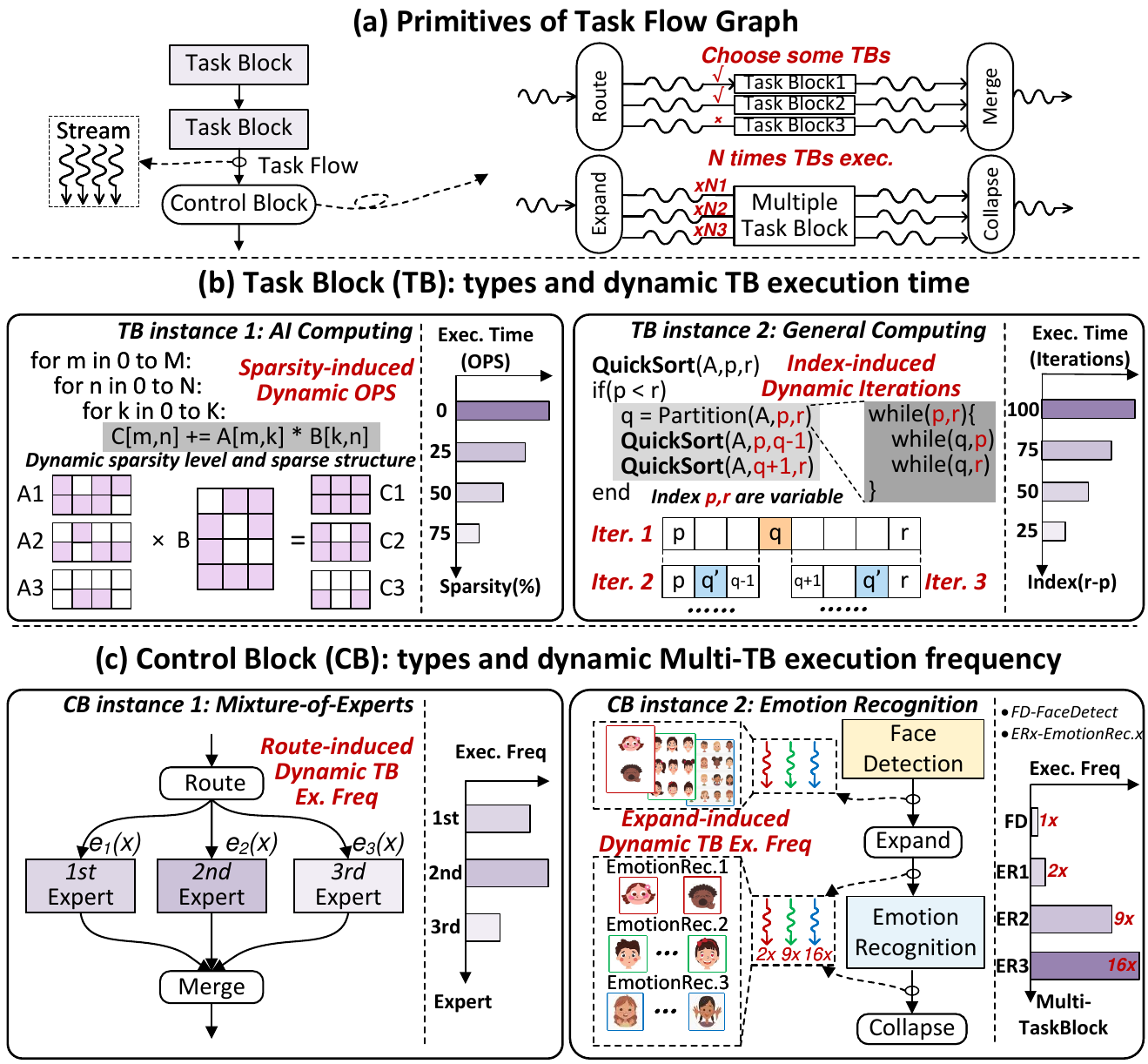}
    \caption{Orchestrated Workflow Graph and Dual Dynamicity.}
    \label{figback}
\vspace{-20pt}      
\end{figure}

\subsection{Orchestrated Workflow Graph}

The Orchestrated Workflow Graph (OWG) is defined to\textit{ articulate the intrinsic logical correlations and inherent features} of Orchestrated AI Workflows. We will illustrate the OWG primitives in the context of Emotion Recognition and reveal its Dual Dynamicity, as shown in the lower part of Figure \ref{fig1}. 

In the OWG, the Orchestrated AI Workflow execution logic is represented by graphs with nodes and edges, where the edges are \textit{Task Flows}, and nodes are \textit{Task Blocks (TBs)} and \textit{Control Blocks (CBs)}, as shown in Figure \ref{figback} (a).

\textbf{Task Flow:} Orchestrated Workflow encapsulates the dataflow, delineating the data transmission relationships between TBs. 
Task Flow consists of several streams, each consisting of several sequenced data packets and a meta-segment for identification and tagging.
For instance, in the assumed Emotion Recognition scenario, a single port (Task Flow) in a video scene gets multiple videos (streams), each with multiple data frames (data packets).

\textbf{Task Block:} 
In Orchestrated AI Workflows, Task Blocks (TBs) serve as data processing units capable of handling AI Models and general-purpose algorithms, encapsulating various processing steps to generate outputs. As shown in Figure \ref{figback} (b), TBs experience dynamic fluctuations in execution times due to computational loads and memory demands specific to the input data. This variability stems from Sparsity-induced Dynamicity in AI Computing, where sparsity levels in model input activations differ across layers based on input features, impacting execution times. For General Computing tasks, variability derives from Index-induced Dynamicity, as seen in algorithms like QuickSort, where the initial input sequence arrangement affects computational loops and memory requirements, thus altering execution times dynamically.

\textbf{Control Block:} 
Control Blocks (CBs) utilize two pairs of control processes: Route and Merge, and Expand and Collapse. The Route and Merge processes channel the stream into different Task Flows, while Expand and Collapse split the stream into substreams for a single Task Flow. \textit{The dynamic nature of CBs significantly influences the execution frequency of multiple TBs, as the characteristics of the input stream are key in determining the frequency and selection of the Task Flows processed by these TBs.}

\textit{1. Route and Merge:}
The Route Block directs the input stream into selected Task Flows for processing, with their outputs then unified by the Merge Block. Illustrated in Figure \ref{figback} (c), a Mixture of Experts (MoE) model is integrated into a Router Block, distributing an input token across multiple downstream Task Flows, each corresponding to a different expert model. The results from these diverse expert models are consolidated by the Merge Block. The activation of the expert models fluctuates with each token series, causing variations in the execution frequencies of the related Task Flows. These frequencies dynamically adapt to the inputs, a phenomenon known as Route-induced Dynamicity.

\textit{2. Expand and Collapse:} 
The Expand Block splits the input stream into several substreams, each processed through the same Task Flow, while the Collapse Block reassembles these Substreams into a unified stream. As depicted on the right side of Figure \ref{figback} (c), this process is demonstrated in an Emotion Recognition scenario where an input image with multiple faces is divided by the Expand Block. Each face results in a separate substream processed through a sequence of Task Blocks for emotion analysis. The Collapse Block then combines these processed faces back into a single image stream. The variable number of faces in different images leads to a fluctuating number of substreams, dynamically altering the execution frequency of the involved TBs. This variability introduces what is termed Expand-induced Dynamicity.

\noindent\textbf{Dual Dynamicity of Orchestrated AI Workflows:} 
(1) The Control Block dynamically adjusts \textit{the frequency of the subsequent Task Flows}, which in turn alters \textit{the execution frequency of multiple Task Blocks (Multi-TBs)} through which these Task Flows traverse. (2) \textit{The execution time of certain Task Blocks can dynamically vary} in response to changes in the volume and characteristics of the input data. 


Orchestrated AI Workflows stands out from conventional AI by featuring "Dual Dynamicity". This trait requires dynamic reallocation of hardware resources based on varying input data, as opposed to traditional AI, which remains unchanged in its hardware structure regardless of input variability. This feature boosts the efficiency of Orchestrated AI Workflows and affects their hardware requirements.

\section{Spatial Architecture and Challenges}
\label{motiv}

Spatial architecture holds significant potential for Orchestrated AI Workflow due to its reconfigurability, which exposes low-level computation and network resources to the hardware/software interface. This allows reallocating resources for tasks like computing, decision-making, and AI computations, with enhanced pipeline parallelism. In this section, we delve into the management of spatial architecture in Orchestrated AI Workflow, highlighting the necessity for innovative solutions specifically designed to meet the unique challenges and constraints imposed by its Dual Dynamicity.

\subsection{Spatial Architecture}

Figure \ref{fig2} presents a basic Processing Element Array (PEA) within a Spatial Architecture (SA). The core unit, the Processing Element (PE), features an Arithmetic Logic Unit (ALU), a register file, and a configuration buffer, designed for executing algorithmic operators needed for both general and AI computations. These PEs, connected to facilitate pipelined data transfer through network switches at the array's edge, interact with external arrays. Additionally, the SA includes multiple PEAs with integrated memory resources, all managed by a central control core that supplies configuration details. These PEAs form \textit{PEA Clusters}, each specialized to handle specific algorithms efficiently, enhancing the overall workflow execution.


\begin{figure}[t]
    \setlength{\abovecaptionskip}{0pt}
    \centering
    \includegraphics[width=\linewidth]{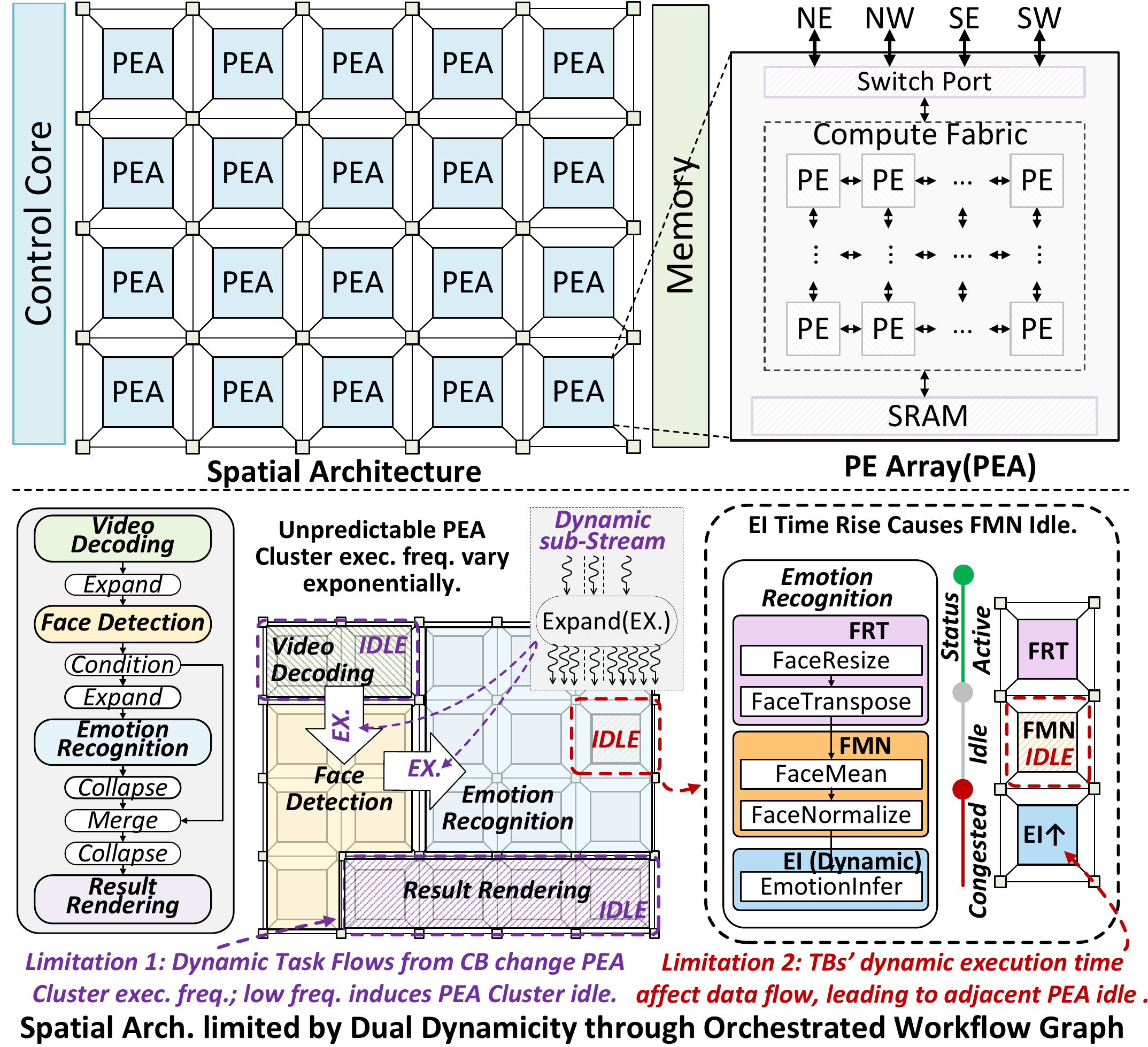}
    \caption{Basic Spatial Architecture and Limitations in Handling Orchestrated AI Workflows.}
    \label{fig2}
    \vspace{-20pt}
\end{figure}


Conventional spatial architecture predominantly uses a static mapping strategy during Orchestrated AI Workflows \cite{prabhakar2017plasticine,swamy2020taurus,tenstorrent,vilim2020gorgon,swamy2022taurus,zhang2021sara,aberger2017emptyheaded,ko2022accelerating,weng2020hybrid,scheduling_zb5,li2022scheduling_zb7}, assigning each TB a predetermined number of PEAs to form a detailed spatial pipeline within these PEAs to execute OWGs. This method facilitates the execution of OWG pipelines at the PEA level. 
However, the Dual Dynamicity of OWGs introduce a significant challenge to the static mapping method employed in conventional spatial architectures, as they tend to cause load imbalance. As a consequence, two important observations can be made:
\textit{(1) The dynamic execution time of a TB affects the data flow in the PEA, affecting the idle and congested states of the input and output loads, leaving preceding or subsequent PEAs idle.}
\textit{(2) The fluctuation in execution frequency of Multi-TBs significantly affects idleness levels within a PEA Cluster incorporating several PEAs.}
These limitations contribute to a significant reduction in utilization.
Previously, some work \cite{tan2022drips,Fifer} somehow noticed the load imbalance and the irregular pipeline problem of the spatial architectures, but cannot cope with the two-level dual dynamicity of OWG. Details are in Section \ref{sec:rel}.



\subsection{Spatial Architecture Challenges} 

The current spatial architecture imposes limitations on the processing of Orchestrated AI Workflows. To address this, dynamic scheduling becomes imperative. This leads us to a thorough examination of the challenges that current spatial architectures encounter in implementing an effective dynamic schedule during the processing of Orchestrated AI Workflows.

\begin{figure}[t]
    \setlength{\abovecaptionskip}{0pt}
    \centering
    \includegraphics[width=\linewidth]{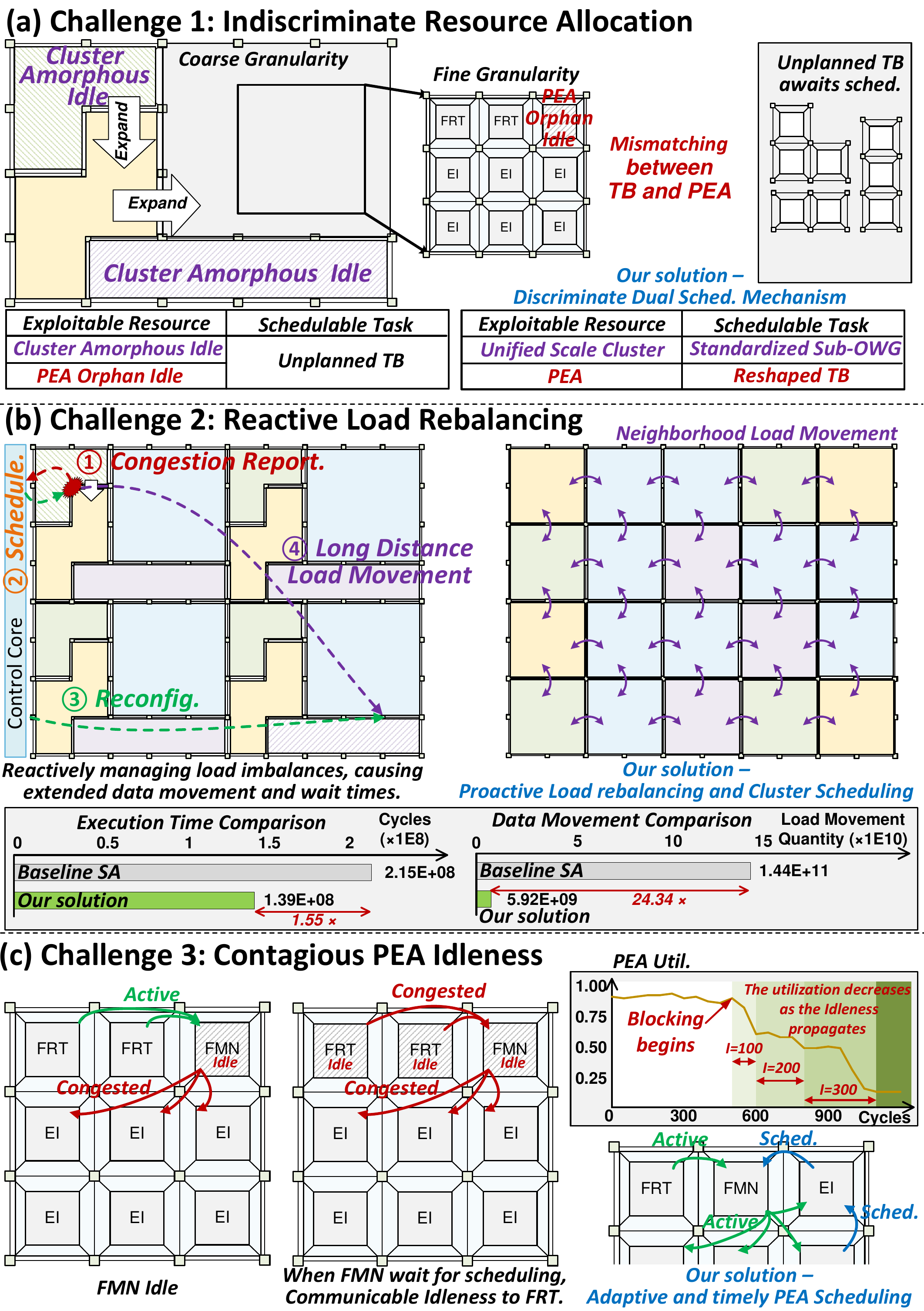}
    \caption{Challenges Issues faced by present spatial architectures in executing a dynamic schedule effectively while managing Orchestrated AI Workflows.}
    \label{fig3}
    \vspace{-20pt}
\end{figure}



\noindent\textbf{Challenge 1: Indiscriminate Resource Allocation:}

The Indiscriminate Resource Allocation challenge primarily arises from the fact that the current spatial architecture allocates schedulable tasks to two distinctly different types of exploitable PEA resources without differentiation. Dual Dynamicity of Orchestrated AI Workflows leads to the emergence of two distinct PEA idles, namely \textit{Cluster's Amorphous Idle} and \textit{PEA's Orphan Idle}. Scheduling in spatial architectures involves efficiently using the exploitable PEA resources to perform the schedulable tasks. However, the current scheduling approach engages with these diverse exploitable PEA resources without effectively disassociating the Dual Dynamicity. Consequently, unplanned TBs are inherently incapable of adapting to exploitable PEA resources. 


To overcome these issues, a revised discriminate dual-scheduling mechanism is necessary, one that discriminates effectively in addressing the Dual Dynamicity of Orchestrated AI Workflows, and distinctively tailors strategies for dual-granularity idle resources. 
Two separate atomic scheduling primitives should be established at each dynamic layer to coordinate between scheduling tasks and resources. 
For PEA Clusters, a unified scheduling shape should be marked to eliminate its amorphous characteristics to achieve seamless task scheduling. For single PEAs, the TB workload maintains corresponding size with PEA configurations through specific operations like partitioning or reshaping, thereby eliminating mismatches in the mapping. By customizing these approaches, scheduling can become more adaptive and efficient, aligning better with the Dual Dynamicity and enhancing overall resource utilization.

\begin{figure*}[!t]
    \setlength{\abovecaptionskip}{0pt}
    \centering
    \includegraphics[width=\linewidth]{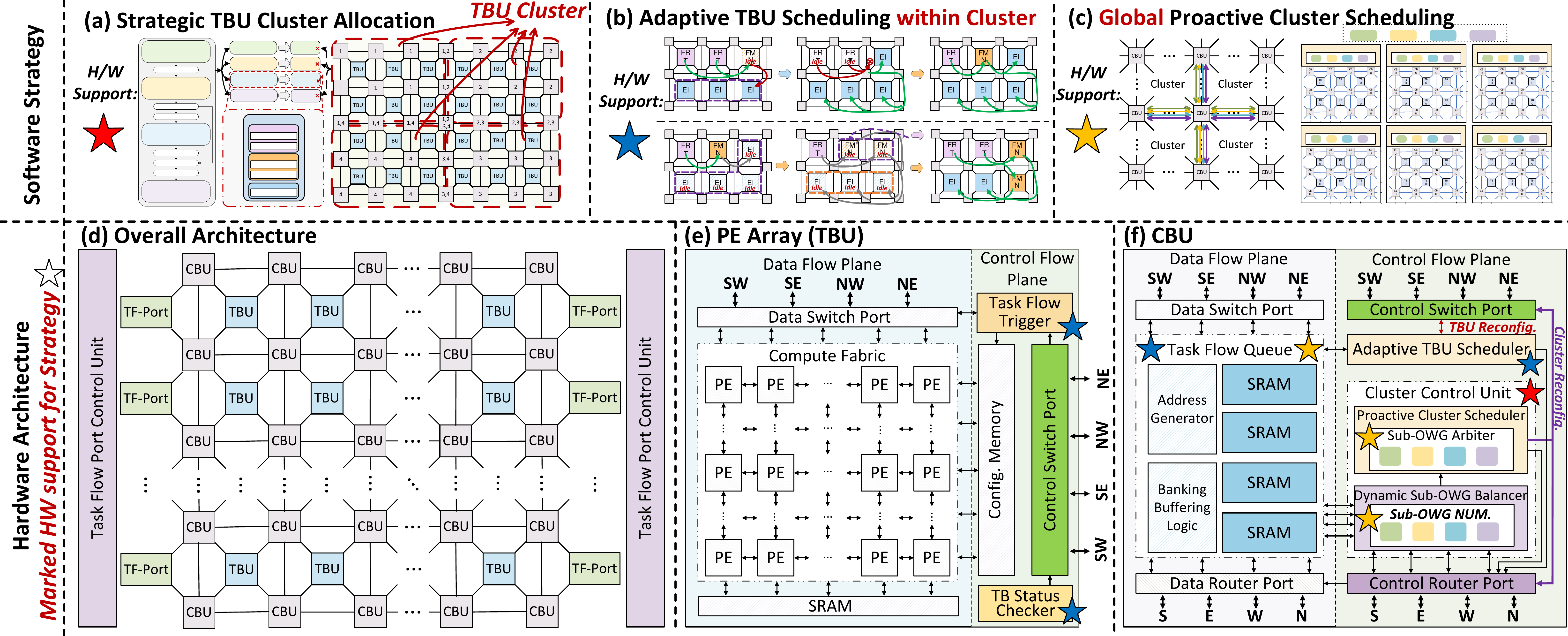}
    \caption{Overall design encompasses a control-flow plane architected spatial architecture, supported by a suite of software strategies.
    Each strategy is backed by the microarchitecture of the control flow plane within TBUs and CBUs.}
    \label{fig4}
    \vspace{-20pt}
\end{figure*}

\noindent\textbf{Challenge 2: Reactive Load Rebalancing: }
Traditional spatial architectures typically employ a reactive scheduling approach in which the system maintains a specific configuration pattern during operation. This process struggles with dynamic scenarios that lead to an accumulation of unbalanced workloads. Once this imbalance reaches a predefined threshold, it triggers a reactive scheduling intervention. At this point, the control core issues reconfiguration and data transfer requests, facilitating resource reconfiguration, and completing the scheduling process. \textit{However, this method does not proactively alleviate workload imbalances, often resulting in prolonged data transfers and waiting periods.} The duration of load rebalancing is explicitly reflected in cluster idleness, highlighting inefficiencies in the process. 

Our proposal introduces a Diffusive Load-Balancing model designed to proactively predict and address workload imbalances, enabling early rebalancing efforts. This model allows each cluster to autonomously adjust tasks and data based on local workload changes, fostering decentralized scheduling. With anticipatory adjustments, the system pre-prepares and optimizes necessary tasks and data, minimizing data movement and idle times, thus boosting system efficiency. In our experimental evaluations, particularly with Crowd Mask Recognition, we noted that the baseline SA with reactive load rebalancing incurred 1.55 times more idleness and 24.34 times the amount of load movement than our proactive strategy.

\noindent\textbf{Challenge 3: Contagious PEA Idleness: }
In traditional spatial architecture, PEAs (Processing Element Arrays) are interconnected, forming a spatial pipeline for continuous process flow. Each PEA executing at variable times can create load imbalances, leading to some PEAs idling due to either depleted input data or congested output channels, thereby generating pipeline bottlenecks. Without swift configuration adjustments, this idle state may spread, affecting both upstream and downstream PEAs. Stage 4 of One-Shot Video Object Segmentation was studied to analyze the impact of Contagious PEA Idleness without scheduling adjustments, as shown in Figure \ref{fig4} (c), indicating a trend of reduced utilization. Depending heavily on a central control core to detect and mitigate these bottlenecks in systems with numerous PEAs is highly inefficient.

Therefore, our insight suggests the development of an adaptive, self-scheduling capability for PEAs, combined with the formation of a semantic layer for control information transmission between PEAs. With this approach, idle PEAs would proactively request tasks from neighboring, congested PEAs, enabling prompt rescheduling. This proactive approach not only prevents the spread of the contagiously idle state by maintaining operational continuity in real-time but also enhances the overall efficiency of the spatial pipeline. By shifting from a centralized detection and management model to this more distributed and responsive structure, spatial architectures can better accommodate dynamic workloads and prevent Contagious PEA Idleness.

\section{Design}\label{sec:des}

The existing spatial architecture reveals limitations in deploying the Orchestrated AI Workflow due to its Dual Dynamicity, highlighting challenges in Indiscriminate Resource Allocation, Reactive Load Rebalancing and Contagious PEA Idleness.

We introduce our design, Octopus, an innovative scale-out spatial architecture and scheduling strategies tailored to the effective deployment of Orchestrated AI Workflow. As shown in Figure \ref{fig4}, Octopus highlights a proposed scheduling approach that incorporates \textit{Discriminate Dual-Scheduling Mechanism}, \textit{Adaptive TBU Scheduling Strategy}, and \textit{Proactive Cluster Scheduling Strategy}. Additionally, the control flow plane \cite{2023deng} of Octopus components is fully architected to support these strategies. 



\subsection{Overall Architecture}\label{sec:4.1}

Octopus is a tiled scalable architecture tailored to the effective deployment of Orchestrated Workflow Graph, including the reconfigurable Task Block Processing Unit (TBU) and Control Block Processing Unit (CBU), as shown in Figure \ref{fig4} (d). Octopus is designed to segment and map various parts of the OWG onto each of its components. The control flow plane serves as the central component of Octopus's operation, which encompasses the intrinsic control flow planes within TBUs and CBUs, as well as a specialized Control Network dedicated to the control information management. This control information is essential, as it orchestrates the data flow plane activities to ensure the proper execution of data processing tasks. Octopus uses the Task Flow Port to manage and simplify data transfer with DRAM.

\textbf{TBU:} 
The TBU of Octopus has been enhanced with a novel control flow plane, designed to facilitate autonomous status feedback and enable agile scheduling responses. We have decoupled the traditional control flow components from the PEA, an optimization depicted in Figure \ref{fig4} (e), showing the architecture of the Octopus TBU. While the traditional computing functions remain in the data flow plane, the control flow plane now features three new related micro-architectures: the Task Block Status Checker (TB Status Checker), the Control Switch Port, and the Task Flow Trigger (TF Trigger). 
The TB Status Checker monitors the TB's status and sends configuration requests via the Control Switch Port to CBUs. The CBUs then send control info to the TBU, causing the TF Trigger to initiate the data flow plane for new configuration executions.

\textbf{CBU:} 
The CBU integrates distributed memory and router structure, and importantly, it serves as a pivotal component of decentralized control, as illustrated in Figure \ref{fig4} (f). 
The CBU's data flow plane includes Data Switch Port, Data Router Port, Control Block Engine (CB Engine), and Task Flow Queue (TF Queue) based on a scratchpad.
CB Engine is responsible for managing the Route, Merge, Expand, and Collapse control processes.
TF Queue has multi-bank SRAM and control logic for storing pipeline results between TBUs and facilitating data transfer with nearby switches.
CBU's control flow plane integrates an Adaptive TBU Scheduler, Cluster Control Unit (Cluster CU), and Control Ports facilitating communication with other CBUs and TBUs. 
The detailed functionalities of these components will be elaborated upon in Sections \ref{sec:4.3} and \ref{sec:4.4}. 

\textbf{Interconnect:}
The Octopus consists of two separate networks: Control Network and Data Network. Both networks share identical topological frameworks. Control Network, although low-bandwidth, suits our scheduling strategy.

\textbf{Decentralized control in Octopus:}
The control flow plane, established as the Octopus architecture, plays a crucial role as the hardware foundation that allows dynamic scheduling. This is primarily because it shifts control mechanisms down to the CBU and TBU levels, moving away from the previously centralized control paradigm. CBUs, following the OWG, configure the connected TBUs sequentially and convey this configuration onward to subsequent CBUs. This process establishes a spatial pipeline at the TBU level through the relay of control information. The transition to decentralized control endows CBUs and TBUs with the capacity to monitor and respond to load conditions in real-time with heightened efficacy. Through this approach, they are able to dynamically amend control information, thus promoting agile and autonomous scheduling. 

\subsection{Discriminate Dual-Scheduling Mechanism}\label{sec:4.2}

\begin{figure}[t]

    \setlength{\abovecaptionskip}{0pt}
    \centering
    \includegraphics[width=\linewidth]{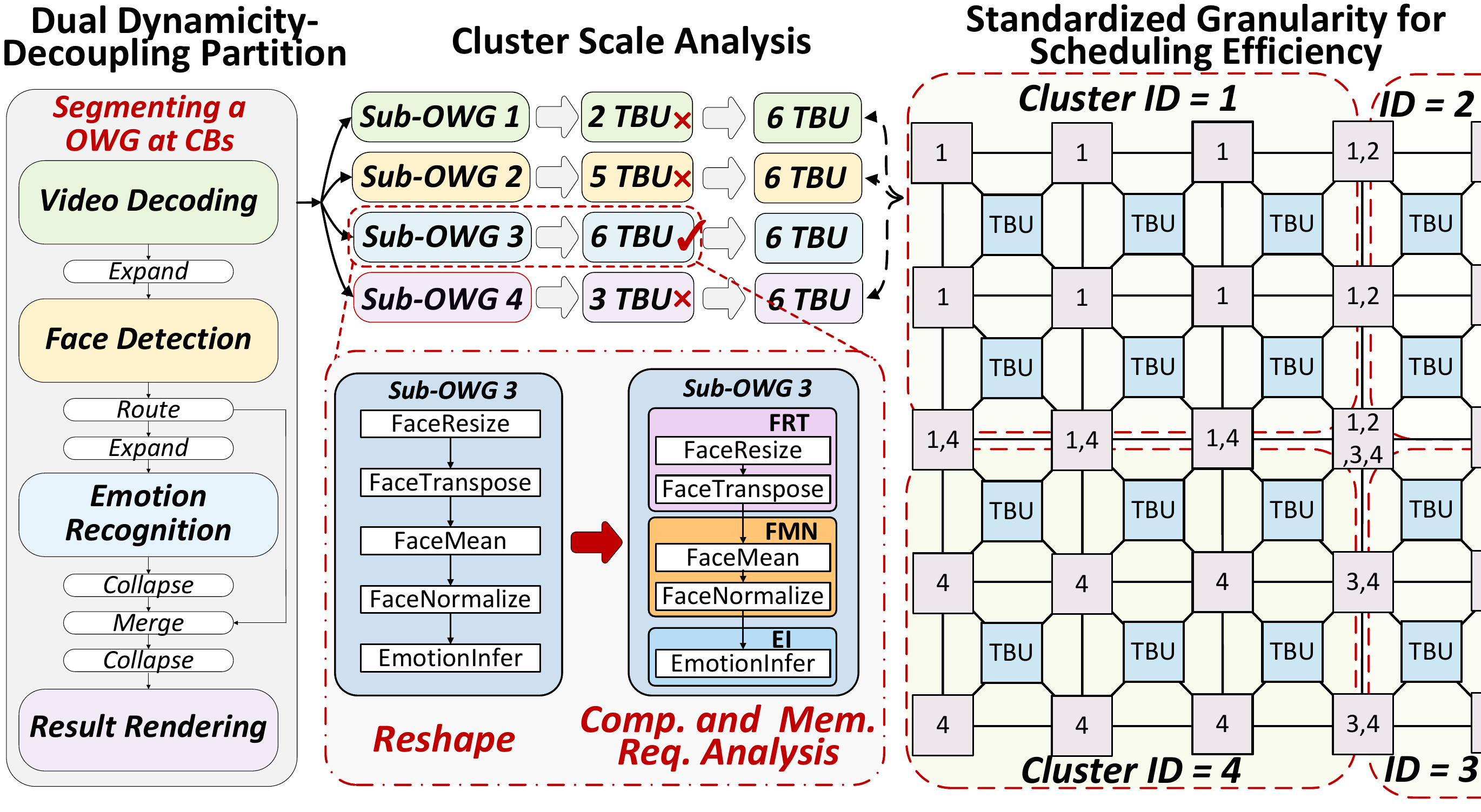}
    \caption{Discriminate Dual-Scheduling Mechanism.
    }
    \label{fig5}  
    \vspace{-20pt}
\end{figure}


To address the challenge of Indiscriminate Resource Allocation, the Discriminate Dual-Scheduling Mechanism has been proposed as a means to establish two separate atomic scheduling primitives at each dynamic layer to coordinate between scheduling tasks and resources. This section explores key questions about how Discriminate Dual-Scheduling Mechanism affects the cluster size and management of TBU Cluster using the CBU Cluster Control Unit.

\textbf{Key insight:}
The challenges of Indiscriminate Resource Allocation underscore the importance of the Discriminate Dual-Scheduling Mechanism, which effectively addresses resource and task allocation through stratified granularity provoked by Dual Dynamicity. This mechanism establishes two separate atomic scheduling primitives between tasks and resources at two levels of dynamic granularity. By partitioning the OWG into sub-OWGs at the CB, we identify two scales of schedulable tasks: sub-OWG and TB. Concurrently, we standardize the scale and shape of clusters, which aligns with two scales of hardware resource availability: Cluster and TBU. In coarse-grained scheduling, the varying execution frequencies of Multi-TB within Dual Dynamicity only impact the scheduling strategies of different clusters for different sub-OWGs, leading to a coarse-grained Proactive Cluster Scheduling. In fine-grained scheduling, variations in TB execution times are confined within individual clusters, affecting only the internal PEA scheduling for executing TB within the current sub-OWG, thereby enabling fine-grained Adaptive TBU Scheduling. In general, this structured approach not only standardizes the scheduling framework, but also enhances agile resource allocation and scheduling across different levels of granularity.

\textbf{Strategy:} 
The process, as shown in Figure \ref{fig5}, begins with the preprocessing of sub-OWGs to tailor them for optimal compatibility with the computational and memory capacities of the TBUs. This step involves reshaping the TBs to maximize PE level resource utilization.
Following preprocessing, the computation and memory characteristics of the sub-OWGs are analyzed to determine the minimal number of TBUs required for efficient operation. Due to the potential variability in workload requirements, estimating the precise number of TBUs necessary for diverse OWG tasks can present challenges.
To ensure seamless scheduling and avoid mismatches across sub-OWGs, we establish the highest TBU requirement observed among them as the standard cluster size for managing the entire OWG workload. Notably, for those sub-OWGs that demand fewer TBUs than the established cluster size, we enhance parallelism within their respective TBs. This adjustment serves to align their TBU usage with the overarching cluster capacity, thereby streamlining data flow and enhancing overall system efficiency.
The Adaptive TBU Scheduling Strategy described in Section \ref{sec:4.3} automatically adjusts TBU numbers for each TB during runtime to ensure a dynamic workload balance in Cluster.

\textbf{Microarchitecture support:} 
Cluster CU assumes a central role in the management of operations related to cluster logic. 
After Discriminate Dual-Scheduling Mechanism partitions, each Cluster CU in the CBUs of the same cluster gets a matching Cluster ID configuration.
The CBU positioned at the forefront of the control flow sequence is designated as the Cluster Leader.
The Cluster Leader is tasked with aggregating statistical data from the sub-OWG input information, which is processed under the Dynamic Sub-OWG Balancer spanning all CBUs within the Cluster, and subsequently orchestrates sub-OWG scheduling.
Further elaboration of these processes is provided in Section \ref{sec:4.4}.

\begin{figure*}[t]
    \setlength{\abovecaptionskip}{0pt}
    \centering
    \includegraphics[width=\linewidth]{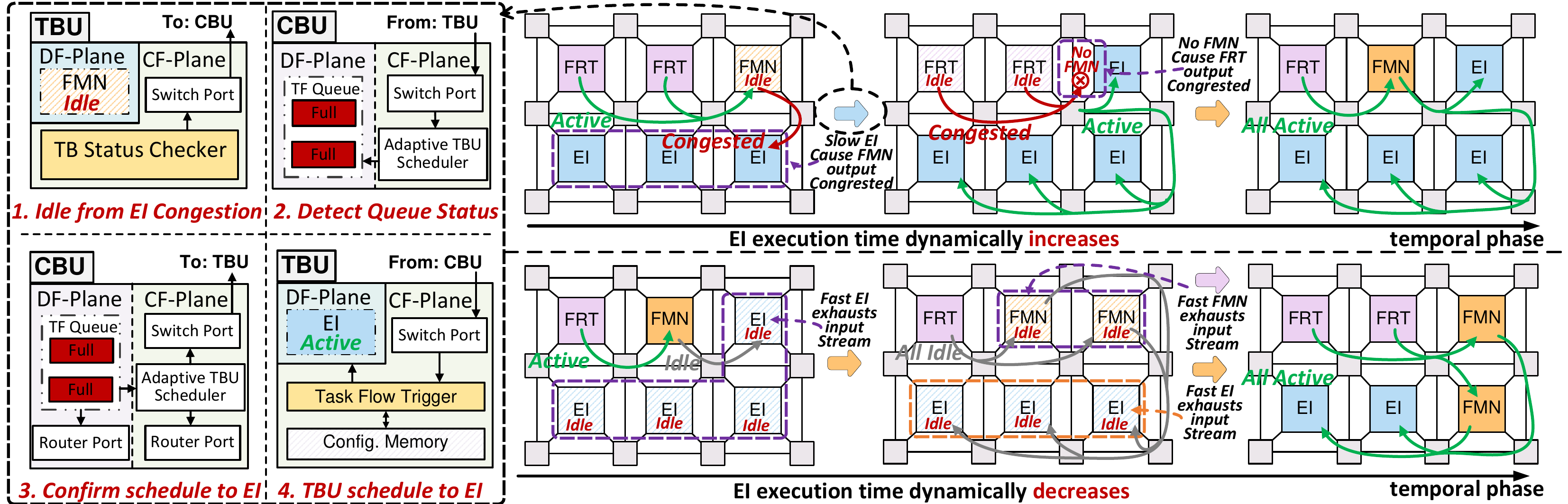}
    \caption{Adaptive TBU Scheduling Strategy. It depicts dynamic scheduling in a Cluster with varying EI execution time (right), and the first scheduling TBU-CBU interaction (left).}
    \label{fig6}
    \vspace{-20pt}
\end{figure*}

\subsection{Adaptive TBU Scheduling Strategy}
\label{sec:4.3}


To address the Contagious PEA Idleness challenge by enabling timely configuration switching of TBUs, Adaptive TBU Scheduling Strategy has been proposed.
This strategy leverages the functionalities of the control flow plane to establish a decentralized communication protocol, facilitating direct communication between TBU and CBU. This enables rapid transmission of control information, configuration switching, and scheduling, thereby resolving the PEA idleness challenge.

This section details the Adaptive TBU Scheduling Strategy's role in fine-grained TB level scheduling within a TBU Cluster, exploring the TBU and CBU microarchitecture interaction for precise scheduling and its adjustment to workload for optimal resource utilization and performance.

\textbf{Strategy:}
The Adaptive TBU Scheduling Strategy tackles the dynamic execution time of TBs by requiring TBU actively monitoring of input and output streams.
This approach allows for a dynamic adjustment to the appropriate TB configuration.
The reasoning is that changes in TB execution times affect how fast the input stream is used and the output stream is produced, which in turn affects the related TBU.
A notable deceleration in the execution speed of the current TBU can result in a significant accumulation of the output stream, produced by preceding TBUs, ultimately leading to a halt. In contrast, delayed generation of the input stream for subsequent TBUs may induce periods of inactivity.

As illustrated in Figure \ref{fig6}, our Adaptive TBU Scheduling Strategy adheres to the following principles: When the TBU input stream is idle or the output stream is congested, the current executing task is halted. In such scenarios, the TBU initiates dynamic scheduling, examines the relevant stream, and switches execution to the corresponding TB. This signifies that the stream's generation and consumption have encountered bottlenecks, necessitating a switch to the appropriate TB to enhance parallelism.

\textbf{Microarchitecture support:} 
The TF Trigger and TB Status Checker in TBU and the Adaptive TBU Scheduler and TF Queue in CBU implement the above strategy.
TB Status Checker is responsible for detecting the execution status of the current TBU.
When the load of the upstream and downstream TBU changes, TB Status detects the impact on the current TBU and sends the idle control information to the CBU.
CBU's Adaptive TBU Scheduler receives CBU control information and detects the corresponding banked queue status in TF Queue.
If the queue status indicates that the upstream and downstream TBs meet the dynamic scheduling conditions, Adaptive TBU Scheduler returns the corresponding TB configuration index to TBU.
The TF Trigger acquires the configuration index and subsequently prompts the TBU to alter the data plane's configuration for the execution of the scheduling procedure.

In consideration of multiple TBUs operating the same TB within the cluster, the Task Flow is allocated and aggregated within the relevant CBUs. Consequently, during the scheduling process, the CBU will also communicates with other CBUs and reconstructs the allocation and aggregation logic. This refactoring considers the reduction of data movement. 

Figure \ref{fig6} illustrates the application of sub-OWG 3 to exemplify the Adaptive TBU Scheduling Strategy and related architecture. EmotionInfer (EI), characterized as a TB with variable execution duration, serves as a model to demonstrate the agility of the scheduling system. We show how the scheduling mechanism dynamically rebalances the workload as the runtime of the EI dynamically increases and decreases, adhering to the above rules.

\begin{figure*}[t]
    \setlength{\abovecaptionskip}{0pt}
    \centering
    \includegraphics[width=\linewidth]{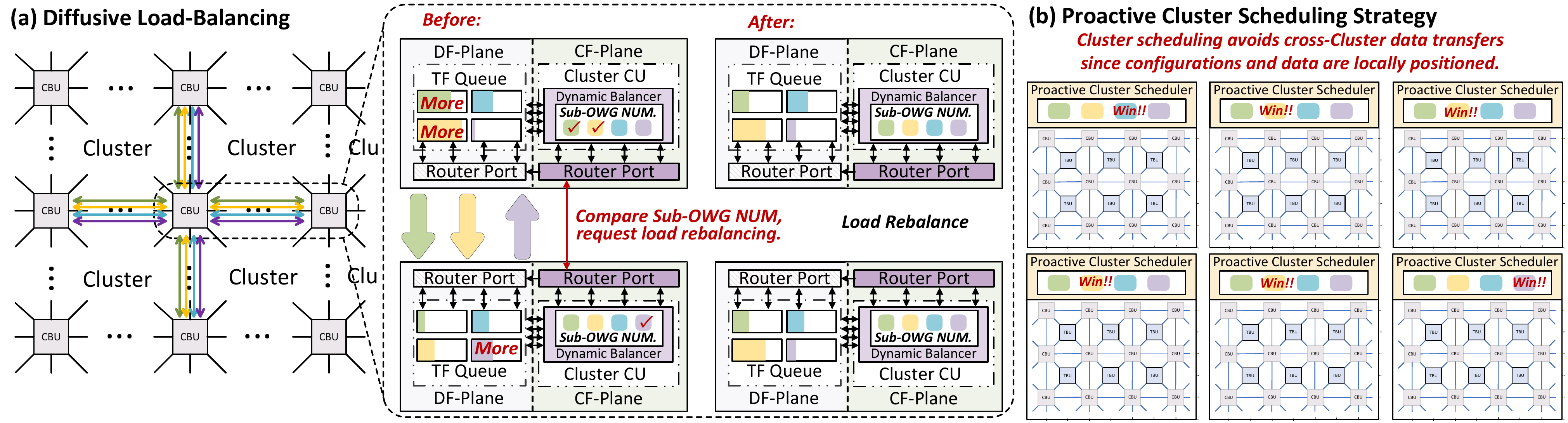}
    \caption{Diffusive Load-Balancing Model and Proactive Cluster Scheduling Strategy.}
    \label{fig7}
    \vspace{-20pt}
\end{figure*}

\subsection{Proactive Cluster Scheduling Strategy}
\label{sec:4.4}


To achieve Diffusive Load-Balancing model and address the Reactive Load Rebalancing challenge, Proactive Cluster Scheduling Strategy has been proposed.
This section examines the Dynamic Sub-OWG Balancer's role in CBU's Diffusive Load-Balancing model across sub-OWGs. We will also review how the Proactive Cluster Scheduling Strategy performs coarse-grained scheduling of sub-OWGs via the Proactive Cluster Scheduler's coordination.


\textbf{Diffusive Load-Balancing Model:}
The Diffusion Load-Balancing Model leverages a distributed framework in which each sub-OWG's input stream is cached in the TF-Queue of a CBU located at a corresponding position within the cluster. These CBUs periodically communicate with the corresponding CBUs of neighboring clusters, compare the accumulated input loads of corresponding sub-OWGs, and rebalance loads according to designated strategies. This proactive and distributed approach to load management helps prevent long-term load accumulation and the need for long-distance data transfers, thus alleviating network congestion. \textit{The periodicity of CBU communication, the volume of load transferred, and the communication range are configurable.} To demonstrate the effectiveness of this model, the simplest strategy was adopted in our evaluation, where each cluster communicates only with its immediate neighbors in the cardinal directions (north, south, east, and west). Load rebalancing occurs when a sub-OWG's load in the current cluster is the largest or second largest compared to its neighbors, targeting the neighbor with the minimum load. Experimental results indicated that the diffusion load balancing model significantly reduces the amount of data needing to be transferred—averaging a reduction by xx times—bringing the system performance close to that of an ideal scheduling scenario, and markedly improving upon traditional centralized load management architectures.

\textbf{Proactive Cluster Scheduling Strategy:} Unlike spatial architectures with centralized control, each Cluster can proactively switch configurations based on its own sub-OWG load conditions. A cluster triggers scheduling when it finds an empty stream, a full output stream, or the tasks executed exceeds a set threshold. Then it compares sub-OWG load accumulation and selects the task from the longest queue for configuration and execution.

\textbf{Microarchitecture support: }The Diffusive Load-Balancing within each CBU is orchestrated by the Dynamic Sub-OWG Balancer. It actively monitors the load of sub-OWGs and communicates with corresponding CBUs in adjacent Clusters, irrespective of whether they hold a leadership role. The primary purpose of Dynamic Sub-OWG Balancer is to facilitate data movement that ensures a balanced load distribution across the network.

The role of the Proactive Cluster Scheduler, managed by the Cluster Leader's CBU, is crucial in the dynamic scheduling of sub-OWGs. It systematically collects load information from the Dynamic Sub-OWG Balancers in all CBUs of the Cluster. This data collection enables the Cluster Leader to utilize the Sub-OWG Arbiter effectively, empowering it to make informed scheduling decisions that optimize overall system performance and maintain load equilibrium consistently.

\section{Implementation}\label{sec:imp}

Here, we discuss the implementation of the software stack, simulator, and hardware used in the evaluation.

\textbf{Hardware:} Our Octopus design is parameterizable (e.g., PE Array size, CBU scratchpad memory size, port widths, etc.) and yields an architectural description shared with the software stack and simulator. 
We opt for a chiplet-based configuration where each TBU includes 1024 PEs, and each CBU is equipped with a 4MB scratchpad. The entire Octopus system comprises 64 TBUs arranged in an 8x8 grid and 81 CBUs organized in a 9x9 grid. In some experiments, we used 4x4 chiplets to package into a wafer-scale architecture.
Table \ref{tab1} shows the hardware parameters. The prototype of Octopus was synthesized using a 7nm TSMC technology library at 1 GHz. 
Our advanced microarchitecture primarily targets the control flow plane and is designed to have a minimal area overhead, thereby ensuring only a slight increase in the total area.

\begin{table}[!b]\normalsize 
\vspace{-17pt}
    \setlength{\abovecaptionskip}{0pt} 
    \caption{Single Chiplet Area breakdown (TSMC 7nm)}
    
    \renewcommand\arraystretch{1.35}
    \resizebox{1\linewidth}{!} 
    {
    \begin{tabular}{llcc}
    \toprule[1.25pt]
    {\textbf{Item}}                                                                           & \multicolumn{2}{l}{\textbf{Component}}            & \textbf{Area ($mm^2$)} \\ \hline
    \multirow{5}{*}{\textbf{\begin{tabular}[c]{@{}l@{}}CBU\\ (43.48\%)\end{tabular}}}          & \multirow{3}{*}{DF-Plane}  & TF-Queue (4MB)     & 3.251                  \\
                                                                                               &                            & Data Port            & 0.030                 \\ 
                                                                                               &                            & CB Engine            & 0.002                  \\ \cline{2-4}
                                                                                               & \multirow{3}{*}{CF-Plane}  & TBU Scheduler        & 0.002                  \\
                                                                                               &                            & Cluster CU           & 0.004                  \\
                                                                                               &                            & Control Port         & 0.0028                  \\ \hline
    \multirow{6}{*}{\textbf{\begin{tabular}[c]{@{}l@{}}TBU\\ (38.58\%)\end{tabular}}}          & \multirow{3}{*}{DF-Plane}  & Comp-Fabric (32x32PEs) & 3.021                  \\
                                                                                               &                            & SRAM                 & 0.660                  \\
                                                                                               &                            & Data Port            & 0.0108                  \\ \cline{2-4} 
                                                                                               & \multirow{3}{*}{CF-Plane}  & TF-Trigger           & 0.0052                  \\
                                                                                               &                            & TB Status Checker    & 0.0004                  \\
                                                                                               &                            & Control Port         & 0.0012
                                                                                            \\ \hline
    \multirow{2}{*}{\textbf{\begin{tabular}[c]{@{}l@{}}Interconnect\\ (15.01\%)\end{tabular}}} & \multicolumn{2}{l}{Data Network}                          & 86.62                 \\
                                                                                               & \multicolumn{2}{l}{Control Network}                       & 5.414                 \\ \hline
    \textbf{\begin{tabular}[c]{@{}l@{}}Memory Port\\ (2.93\%)\end{tabular}}                    & TF Port                    &                      & 17.97                 \\ \hline
    \textbf{Octopus Chiplet}     & \multicolumn{2}{l}{81 CBUs, 64 TBUs, Interconnect, Memory Port}   & 613.34
    \\ 
    \bottomrule[1.25pt]    
    \end{tabular}
    }
    \label{tab1}
    \end{table}

\textbf{Software Stack:} 
Our software stack comprises completely automated procedures for Dual Dynamicity-Decoupling Partition and Strategic TBU Cluster Allocation. The resulting sub-OWGs of this workflow will undergo compilation at TB granularity, a stage that has undergone thorough verification \cite{chin2017cgra}. Utilizing the tagged source code, we generate an LLVM IR. Automated tools then analyze LLVM IRs and generate Control Data Flow Graphs (CDFGs). The final step involves transforming the CDFG into a configuration bitstream according to the hardware model.

\textbf{Simulator:} We have developed a cycle-level accurate simulator. It is parameterizable, allowing for customization of factors. It uses the binary configuration file output from the compiler to verify the functional correctness of the Octopus and to evaluate the performance.

\section{Evaluation Methodology} \label{sec:met}

\subsection{Comparison Methodology} \label{subsec:7.1}

We developed a cycle-level simulator that allows full parameterization of all of its components.
Each implemented strategy can be activated as needed. 
TBU and CBU in our design exhibit a topology mode and a data flow behavior highly reminiscent of PMU and PCU in Plasticine \cite{prabhakar2017plasticine}. Consequently, our baseline architecture selects Plasticine of comparable scale and reconfigured the simulator's interconnection method accordingly.

Initially, we carried out a comprehensive assessment of performance and resource utilization, presenting optimization outcomes for both the Adaptive TBU Scheduling Strategy and the Proactive Cluster Scheduling Strategy individually.
It is important to highlight that baseline architecture does not adhere to a predefined scale limit based on Cluster. Instead, it endeavors to allocate an increased amount of TBU resources to tasks with elevated computational demands.





To demonstrate the effectiveness of Proactive Cluster scheduling, we built a wafer-scale architecture consisting of 4x4 chiplets. This setup was designed to fully demonstrate the capabilities of the diffusion load balancing model and also confirmed that our operational paradigm for Orchestrated AI Workflows is effective in wafer-scale architectures.

Finally, we conduct a thorough comparison of the effects of choosing cluster size and cluster scheduling intervals on overall performance.

\begin{table}[!t]\footnotesize
\setlength{\abovecaptionskip}{0pt} 
  \centering
  \caption{Benchmark Characteristics (32-bit data types)}

\renewcommand\arraystretch{1.2}
\resizebox{1\linewidth}{!} 
{
  \begin{tabularx}{\columnwidth}{l|>{\centering\arraybackslash}m{0.7cm}>{\centering\arraybackslash}m{0.7cm}>{\centering\arraybackslash}m{0.7cm}>{\centering\arraybackslash}m{0.7cm}|>{\centering\arraybackslash}m{0.7cm}>{\centering\arraybackslash}X}
    \toprule[1.25pt]
    Item & ER & DPSR & SFR & CMR & OSVOS & OCR \\
    \hline
    Expand & 4 & 4 & 4 & 4 & 2 & 2 \\
    Condition & 1 & 1 & 0 & 1 & 0 & 0 \\
    sub-OWG & 4 & 4 & 4 & 5 & 4 & 3 \\
    Task Block & 10 & 9 & 7 & 10 & 6 & 10 \\
    \hline
    Cluster Scale & 8 & 4 & 8 & 8 & 4 & 4 \\
    \hline
    Input Type & \multicolumn{4}{c|}{Video} & \multicolumn{2}{c}{Image} \\
    \bottomrule[1.25pt]
  \end{tabularx}
}
\label{tab2}
\vspace{-20pt}
\end{table}

\subsection{Benchmark} \label{subsec:7.2} 

We have chosen Orchestrated AI Workflows developed by various developers in real-world scenarios using the ModelArts development platform \cite{ModelArts}. The workflows have been simplified to make them achievable in our Octopus. ModelArts is an all-in-one AI development platform designed for developers to rapidly create and deploy models, managing the entire AI workflow lifecycle. Our evaluation benchmarks are Emotion Recognition (ER), Driver and Passenger Status Recognition (DPSR), Street Flow Recognition (SFR), Crowd Mask Recognition (CMR), One-Shot Video Object Segmentation (OSVOS), Optical Character Recognition (OCR) \cite{ModelArts} with the relevant parameters outlined in Table \ref{tab2}.

In addition to the basic information in Table \ref{tab2}, ER, DPSR, SFR, and CMR specialize in video processing with resolutions of 1920×1080, 960×720, and 640×380 at 30fps, lasting 3 to 30 minutes. Video benchmarks include 20-50 videos of varying resolutions and lengths. OSVOS and OCR are for image processing, using a resolution of
1800 unique images.

\section{Evaluation}\label{sec:ela}

\subsection{Overall Performance} \label{sec:7.1}


Figure \ref{exp1} illustrates the overall normalized performance and the utilization gain achieved by our design, detailing the distinct enhancements provided by the Adaptive TBU Scheduling Strategy and the Proactive Cluster Scheduling Strategy compared to the Plasticine-like baseline.

Our experimental findings indicate that employing the Adaptive TBU Scheduling Strategy in isolation yields only marginal performance improvements. This is attributed to the use of a relatively superior mapping strategy in the static baseline, coupled with the dynamic TB allocation of more TBU resources, which mitigates the impact of the Adaptive TBU Scheduling Strategy. In contrast, integration of the Proactive Cluster Scheduling Strategy leads to a substantial reduction in execution time.

The combined application of both scheduling strategies results in optimal performance, demonstrating a maximum speedup of 4.26× and an average speedup of 2.50×, with an average utilization of 83.5\%. Notably, even in the most challenging scenarios, such as the SFR case, a significant speedup of 1.77× is achieved.

\begin{figure}[t]
    \setlength{\abovecaptionskip}{0pt}
    \centering
    \includegraphics[width=\linewidth]{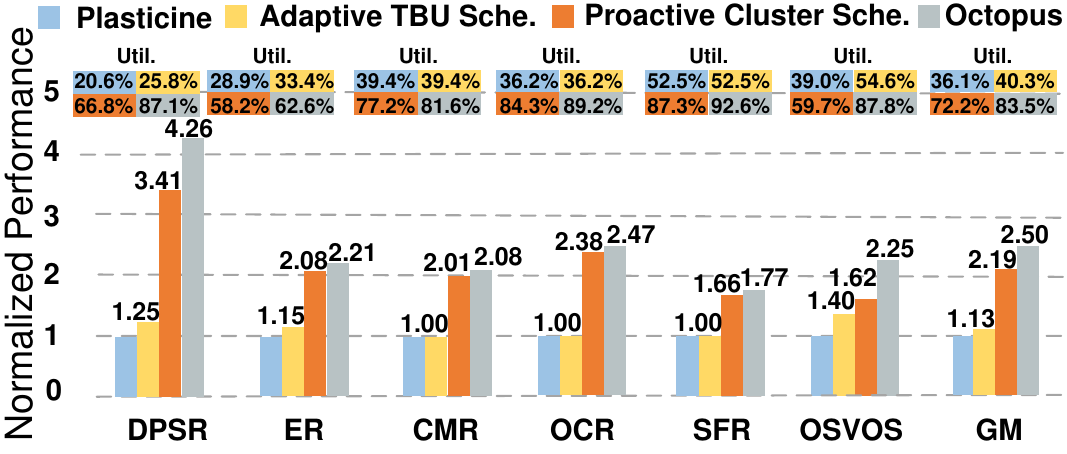}
    \caption{Overall Normalized Performance. Octopus outperforms Plasticine by geomean 2.50×}
    \label{exp1}
    \vspace{-15pt}
\end{figure}

\begin{figure}[t]
    \setlength{\abovecaptionskip}{0pt}
    \centering
    \includegraphics[width=\linewidth]{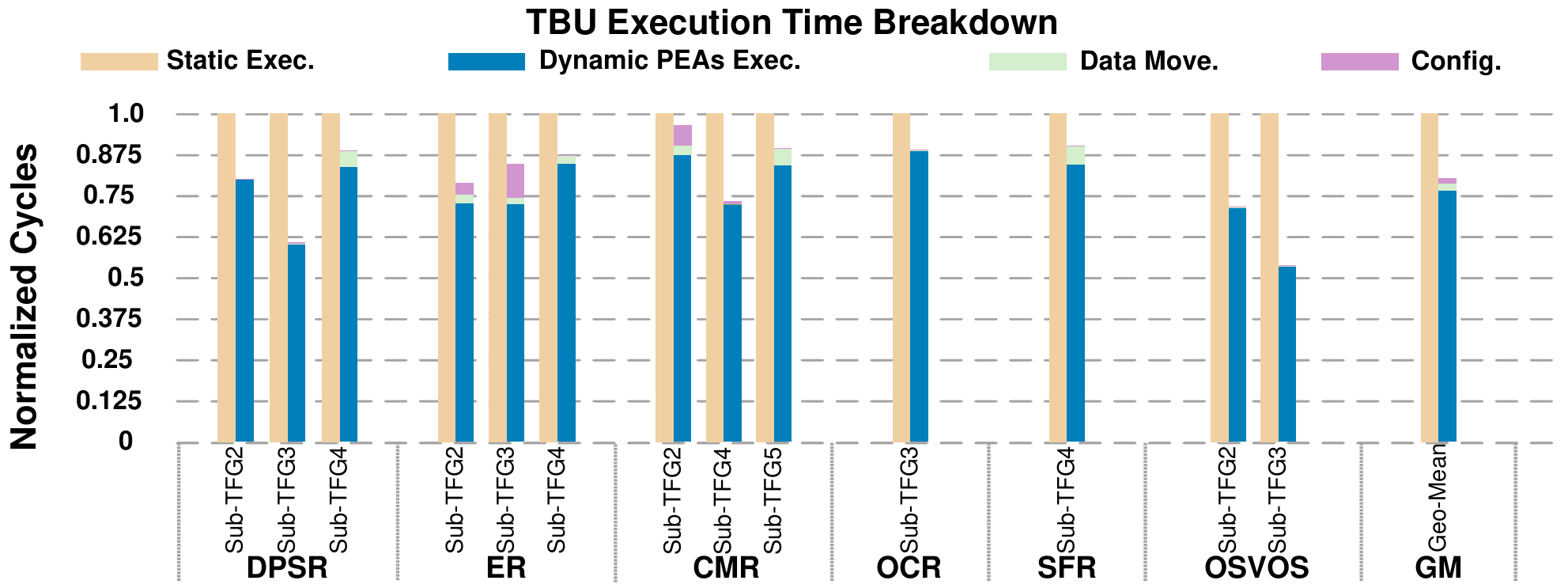}
    \caption{The advantage of Adaptive TBU Scheduling Strategies detailed by normalized breakdown of cycles spent executing each benchmark.}
    \label{exp2}
    \vspace{-20pt}
\end{figure}

\subsection{Time Breakdown Analysis} \label{sec:7.2}

To delve deeper into the enhancements realized by Octopus, we examine the detailed cycle distribution when Adaptive TBU Scheduling is applied in clusters that solely utilize dynamic TB. 
Figure \ref{exp2} highlights these advantages.
Adaptive TBU Scheduling generally boosts performance by 1.25×, with varying gains across benchmarks.
Compute-intensive benchmarks experience greater performance gains, with OSVOS achieving an impressive 1.88× improvement. 
This significant boost can be attributed to the minimal data movement overhead. However, some benchmarks show lower performance gains than others. 
In CMR's sub-OWG2 example, performance boosts by just 1.05× due to few dynamic tasks with little variation in this benchmark. 
Therefore, the static baseline itself is highly utilized, leaving very little room for improvement.


\subsection{Cluster Scheduling under Wafer-scale Architecture} \label{sec:7.3}

\begin{wrapfigure}{l}[0cm]{0pt}
\vspace{-25pt}
   \setlength{\abovecaptionskip}{0pt}
    \centering
    \includegraphics[width=0.5\linewidth]{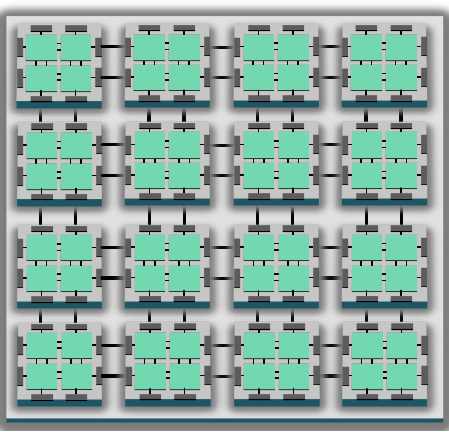}
    \caption{Wafer-scale Arch. Setup.}
    \label{exp44}
   \vspace{-15pt}
\end{wrapfigure}

\noindent\textbf{Experiment setup:}
To fully demonstrate scalability, we expand Octopus to wafer-scale level and construct a set of simulation parameters. These parameters are based on existing commercial and academic Wafer-scale Computing systems\cite{dojo_wafer,Cerebras_WSE,Cerebras_Wafer,UCLA_Wafer}. We scale the transistors to 7nm and assume a frequency of 1GHz for the processor. The number and topology of TBUs and CBUs within each Octopus remain unchanged, and each TBU still has 4MB of SRAM. 16 Octopus Dies are integrated onto a wafer-size interposer, which provides sufficient bandwidth and short enough latency to ensure that the communication capabilities between Dies are similar to those within a single Die. Each Die achieves an external bandwidth of 8TB/s. 3 DIP Cards\cite{dojo_wafer} equipped with HBM2e are plugged into the back of the wafer, providing a total of 96GB DRAM and 1.3TB/s memory access bandwidth. We run benchmarks on this simulation platform with and without our scheduling strategies to demonstrate the improvement of cluster utilization in a large scale.

\begin{figure}[t]
    \setlength{\abovecaptionskip}{0pt}
    \centering
    \includegraphics[width=\linewidth]{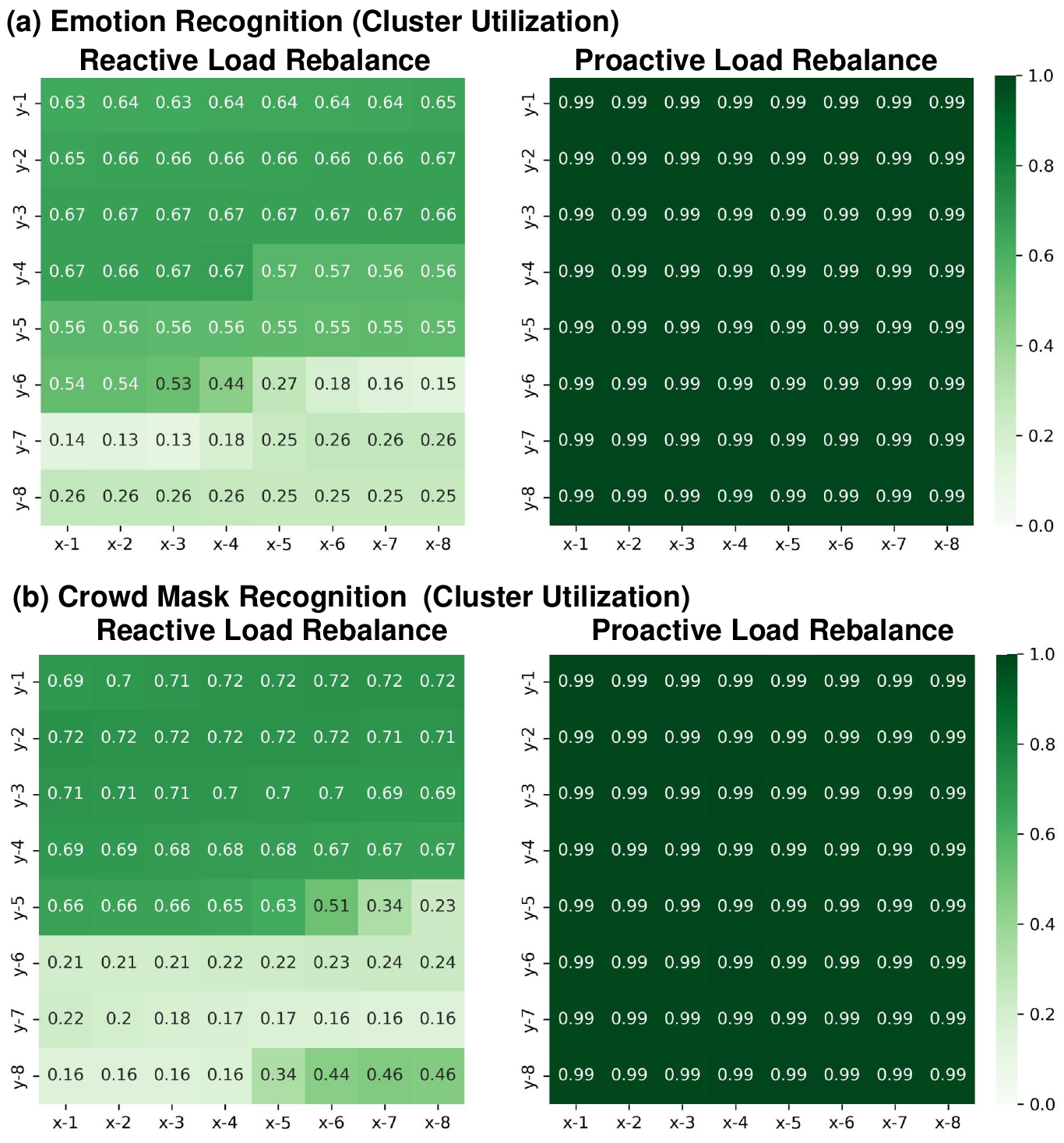}
    \caption{The comparison of proactive diffusion load balancing model and reactive load rebalancing on SA Cluster utilization, taking ER and CMR as examples.}
    \label{exp2.5}
    \vspace{-15pt}
\end{figure}

\begin{figure}[t]
   \setlength{\abovecaptionskip}{0pt}
    \centering
    \includegraphics[width=\linewidth]{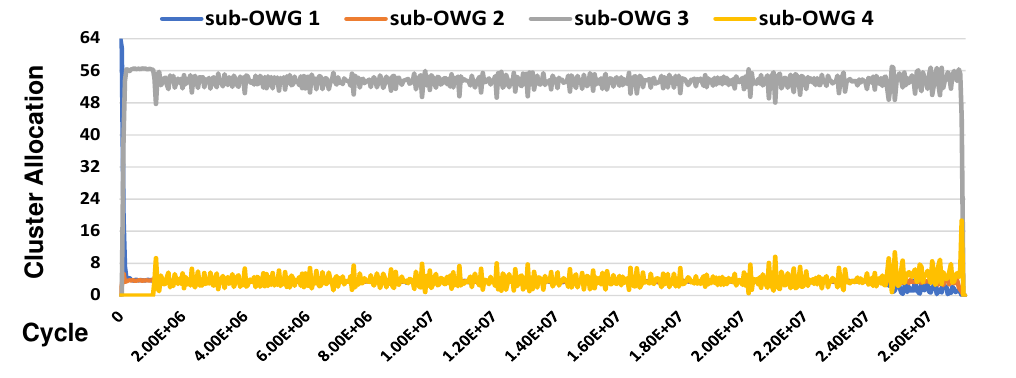}
    \caption{Comprehensive Cluster Scheduling under Octopus, during ER Benchmark Runtime.}
    \label{exp4}
   \vspace{-20pt}
\end{figure}


\noindent\textbf{Evalution of Diffusive Load-Balancing Model:}
We choose ER and CMR as examples within Discriminate Dual-Scheduling Mechanism, to illustrate the effectiveness of the Octopus's proactive Diffusive Load-Balancing Model in wafer-scale. The utilization of each Cluster is visually represented using grayscale images to clearly depict the load balance. In the reactive load-balancing model, periods of idleness are mostly due to delays in load transfer and congestion or idle in the TF Queue. Conversely, the proactive Diffusive Load-Balancing Model experiences minimal idle time, primarily due to brief configuration durations. This comparison highlights the Proactive Cluster scheduling Strategy’s ability to dynamically schedule and distribute loads efficiently across clusters, mitigating delays and enhancing overall system performance.

\begin{figure*}[!t]
    \setlength{\abovecaptionskip}{0pt}
    \centering
    \includegraphics[width=\linewidth]{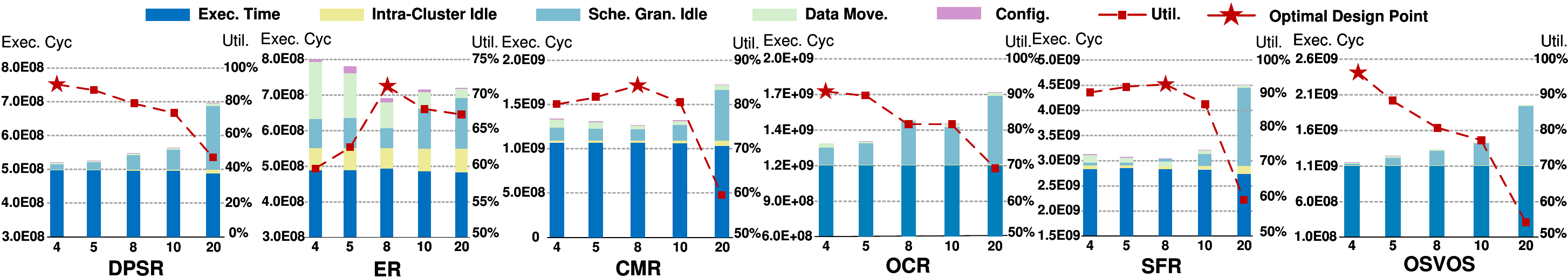}
    \caption{Impact of cluster size on utilization. The optimal cluster sizes for DPSR, ER, CMR, OCR, SFR, and OSVOS are 4, 8, 8, 4, 8, and 4.}
    \label{exp3}
    \vspace{-20pt}
\end{figure*}


\noindent\textbf{Real-time Cluster Scheduling Monitoring:}
To demonstrate how the Octopus operates dynamically at cluster-level granularity during runtime, we present Figure \ref{exp4}, which showcases the runtime TBU allocation of executing the entire ER process on a wafer. To enhance clarity and optimize analysis, we reduced the number of video inputs due to their large size causing extended execution times, which obscured detailed insights in the timing diagrams. Throughout the execution, all clusters are consistently supplied with data of sub-OWG 1. Simultaneously, clusters consume their respective sub-OWG 1s from the queues and generate subsequent sub-OWGs. This dynamic generation of sub-OWGs often trigger reconfigurations to execute other dominant sub-OWGs efficiently. As a result, sub-OWG 3s exhibit the highest allocation ratio during most of the runtime owing to their most frequent generation. As all preceding sub-OWGs are executed, sub-OWG 4s eventually emerge as the dominant task, acquiring a significant portion of the TBU allocation temporarily before being rapidly completed.

\noindent\textbf{Discussion:}
It can be observed that the Discriminate Dual-Scheduling Mechanism demonstrates sufficient scalability to handle wafer-scale scheduling. With adequate bandwidth between chiplets provided by the interposer, the cluster allocation is capable of keeping up with the dynamicity of workloads, greatly enhancing cluster utilization at a large scale.

\subsection{Effect of Cluster Size} \label{sec:7.5}

We noticed that the choice of cluster size has a notable impact on utilization. 
To probe this, we conducted extensive experiments that tested various cluster allocations on all benchmarks, as illustrated in Figure\ref{exp3}. 
The optimal cluster size varies for different workloads due to distinct computing memory access ratios and data dependency characteristics. 
The sensitivity to variations in performance overheads differs among benchmarks due to their distinctive features. 
A smaller cluster size restricts scheduling space at the TBU level and increases data transfer costs, leading to performance degradation. 
On the other hand, a larger cluster size limits cluster-level scheduling, resulting in performance degradation as well.
Based on our evaluation, the optimal cluster sizes for DPSR, ER, CMR, OCR, SFR, and OSVOS are 4, 8, 8, 4, 8, and 4, respectively.
These findings also serve as a foundation for future enhancements to our TBU Cluster Allocation.

\subsection{Effect of Proactive Scheduling Interval}\label{sec:7.6}

\begin{figure}[!t]
    \setlength{\abovecaptionskip}{0pt}
    \centering
    \includegraphics[width=\linewidth]{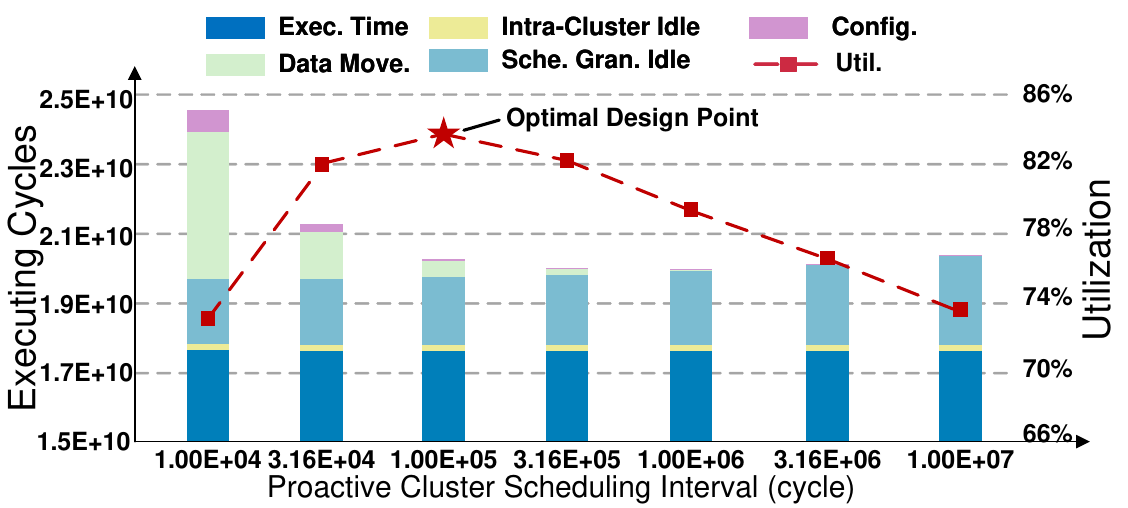}
    \caption{Effect of Proactive Scheduling Interval.}
    \label{exp5}
    \vspace{-20pt}
\end{figure}

To explore the impact of Proactive Scheduling interval on utilization, we conducted a design space exploration using logarithmically distributed scheduling periods. We averaged the execution time for all benchmarks, concurrently calculating utilization for different periods, as shown in Figure \ref{exp5}. 
The long scheduling interval, which leads to load imbalances and similarity to static scheduling, results in minimal idle time in data movement and configuration, as well as low utilization, approximately 73\% at 1.00E+07 cycles. On the contrary, a short interval necessitates frequent data movements and reconfigurations, also reducing utilization to a similar level at 1.00E+04 cycles. Identifying an optimal scheduling period is key; we find that the 1.00E+05 cycles offered a balance between execution time and utilization.

\section{Related Work} \label{sec:rel}

\textbf{Imbalance phenomenon in spatial architecture:}
Studies reveal spatial architecture imbalances, but consistently fail to meet the dynamic scheduling demands of Orchestrated AI Workflows.
Fifer \cite{Fifer} is designed to optimize the efficiency of irregular applications in CGRAs, separating them into a pipeline stage feedforward network. To address larger-scale application scenarios, substantial data movement and cache capabilities are necessary. 
For Orchestrated AI Workflows, significant cache and data transfer features are crucial.
DRIPS \cite{tan2022drips} dynamically rebalances data-dependent streaming application pipelines. However, it suffers from the Task-Resource Mismatch challenge.
Other research delves into finer-grained irregularities, such as non-ideal loop-like control logic. 
Revel \cite{weng2020hybrid} decouples the levels of loops, providing a perspective that aids in decoupled thinking. 
UE-CGRA \cite{torng2021ultra} adjusts voltage frequency for various PEs to reduce irregularities, although it is not always applicable.

\textbf{Exploring CGRAs for Streaming Applications:}
The concept of stream in the Orchestrated Workflow Graph is similar to that of many existing CGRA targets. 
Chordmap \cite{9351547} boosts throughput by enabling concurrent kernel execution with a fixed mapping strategy, similar to \cite{7879212}, but assumes a synchronous data flow graph with constant execution times.
ARENA \cite{9436051} supports CGRA dynamic reconfiguration based on data size, mapping kernels to few configurations, which limits support and usability.
Stream Dataflow ISA \cite{nowatzki2017stream} and associated architectures \cite{infinitystream,streamfloating,Stream-basedmemory,nearstream,taskstream,ovengenstream,DSAGEN,spu,PolyGraph} utilize data streaming to capture tasks with high computing demand marked by long stages, simple control, dependencies, and easy memory access for streaming and reuse.
This concept draws parallels to Task Block acceleration, influencing our formulation of the Orchestrated Workflow Graph and its corresponding description.

\section{Conclusion}\label{sec:con}

Introducing OWGs for Orchestrated AI Workflow analysis, we reveal Dual Dynamicity hindering conventional spatial architectures. To address these, we develop Octopus: a new design that combines spatial architecture with tailored scheduling for Orchestrated AI Workflows. Octopus introduces a proposed scheduling approach that integrates the Discriminate Dual-Scheduling Mechanism, the Adaptive TBU Scheduling Strategy, and the Proactive Cluster Scheduling Strategy. Moreover, the control flow plane within Octopus components is fully designed to facilitate these strategies. Octopus notably boosts performance, surpassing current systems by 2.50× on average, peaking at 4.26× in Orchestrated AI Workflows.

\bibliographystyle{IEEEtran}
\bibliography{references}

\vspace{-30 pt}
\begin{IEEEbiography}[{\includegraphics[width=1in,height=1.25in,clip,keepaspectratio]{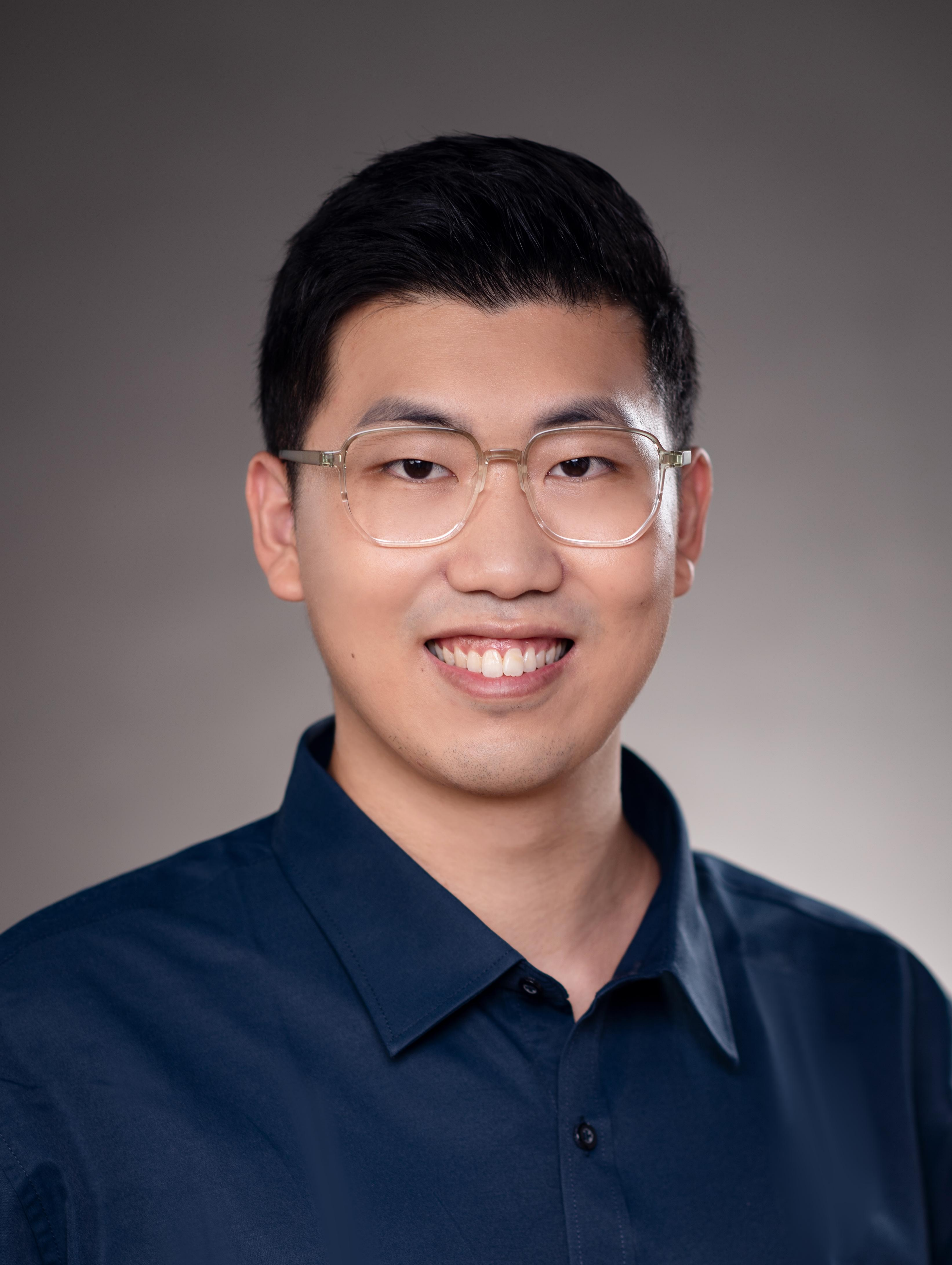}}]{Jinyi Deng}
(Student Member, IEEE) received the B.S. degree in electronic science and technology from the Chien-Shiung Wu College, Southeast University in 2018. He is currently pursuing the Ph.D. degree majoring in Electronic Science and Technology at the School of Integrated Circuits, Tsinghua University. His research interests encompass reconfigurable computing, spatial architecture, and hardware-software co-optimization. 
\end{IEEEbiography}

\vspace{-40 pt}

\begin{IEEEbiography}
[{\includegraphics[width=1in,height=1.25in,clip,keepaspectratio]{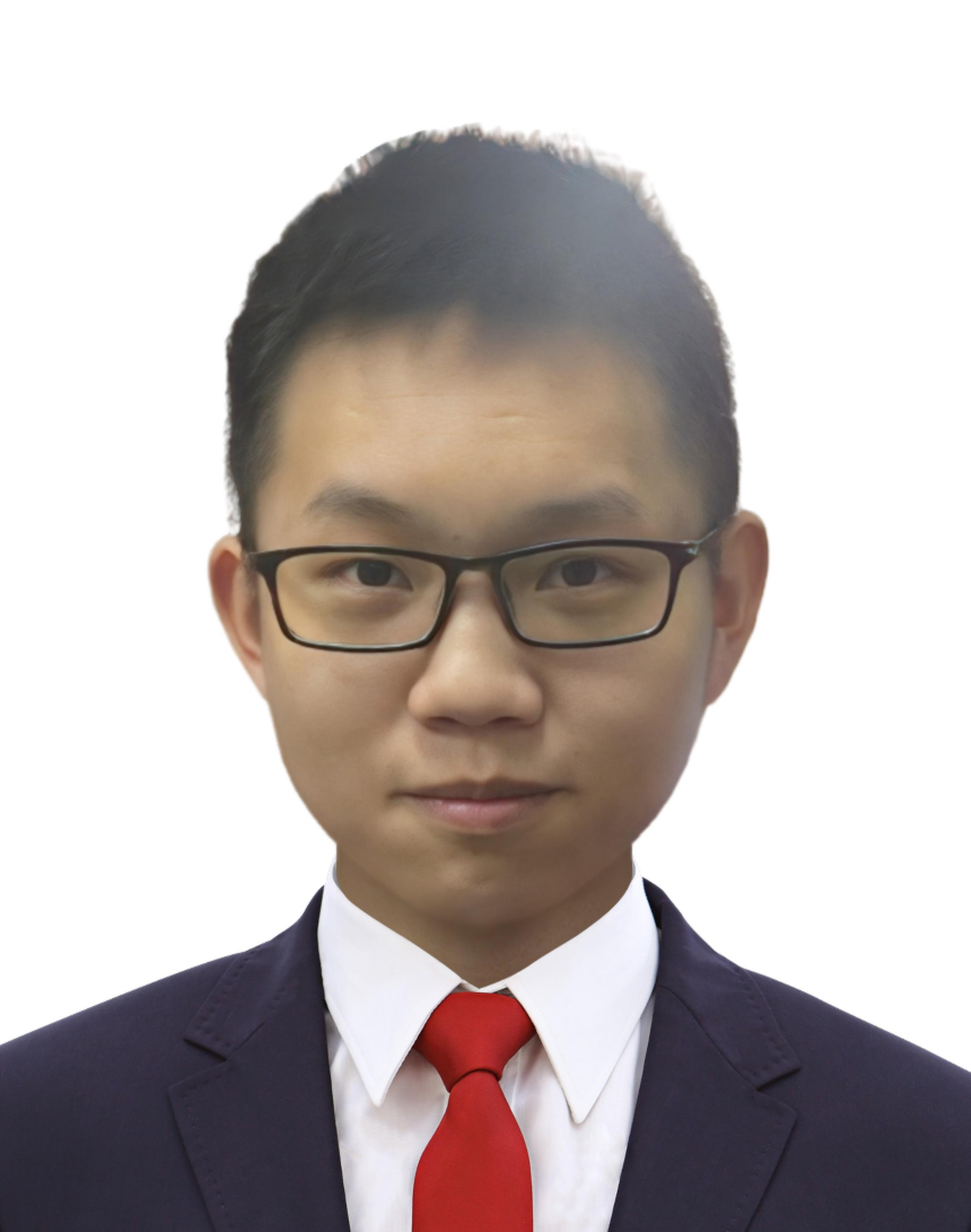}}]{Xinru Tang} received the B.S. degree in Integrated Chip Design and Integrated System from Huazhong University of Science and Technology, Hubei, China, in 2023. He is currently working toward the Ph.D. degree in School of Integrated Circuits, Tsinghua University. His current research interests include microarchitecture of spatial accelerator, AI processor and large-scaling chip design. 
\end{IEEEbiography}

\vspace{-40 pt}

\begin{IEEEbiography}
[{\includegraphics[width=1.17in,height=1.4in,clip,keepaspectratio]{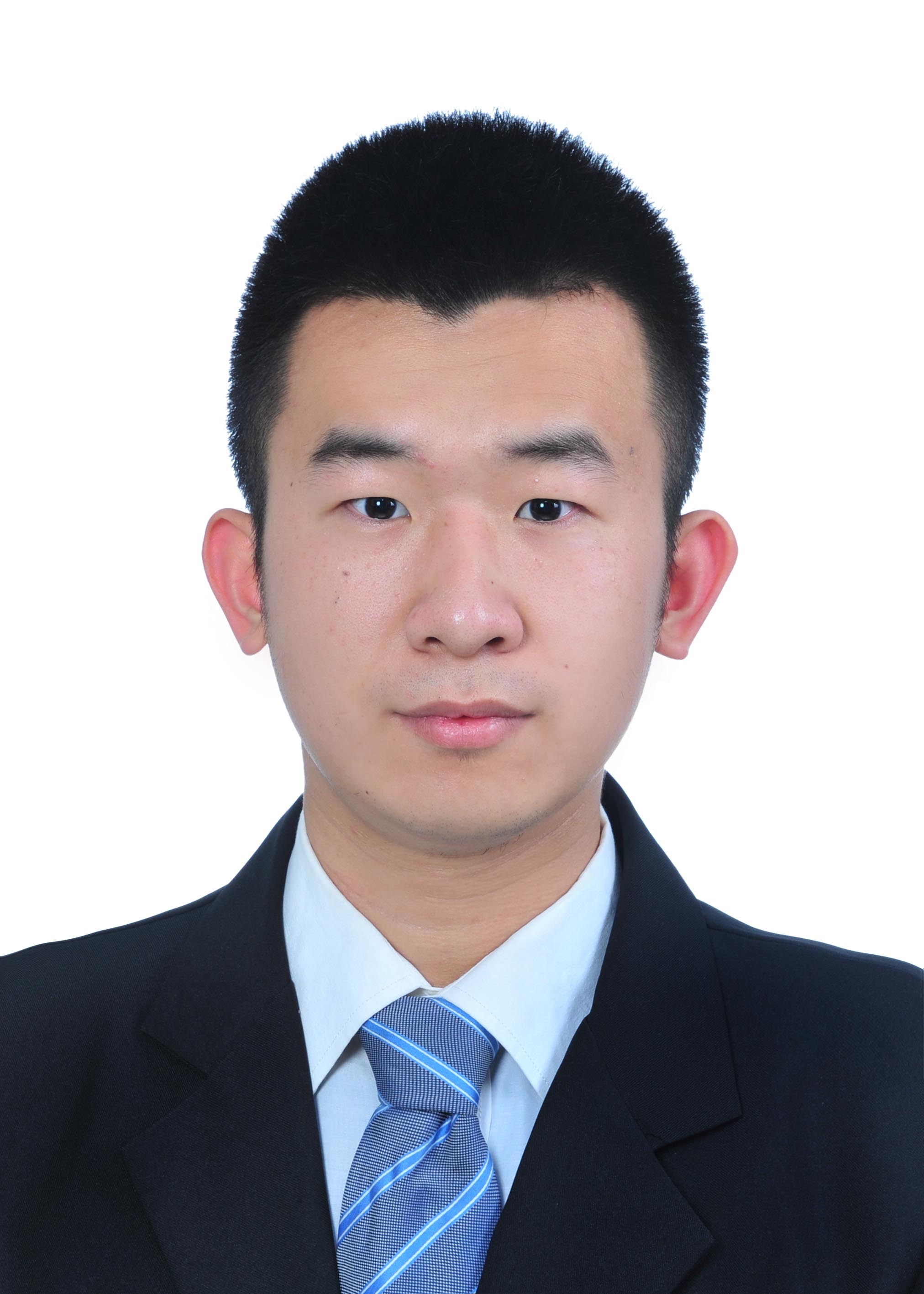}}]{Zhiheng Yue} 
received the B.S. degree in electronic science and technology from the Beijing University of Posts and Telecommunications, Beijing, China, in 2017, and the M.S. degree in electrical and computer engineer from the University of Michigan, Ann Arbor, MI, USA, in 2019. He is currently pursuing the Ph.D. degree in electronic science and technology with Tsinghua University, Beijing.
His current research interests include deep learn- ing, computation-in-memory, AI accelerator, and very-large-scale-integration (VLSI) design.
\end{IEEEbiography}

\vspace{-40 pt}

\begin{IEEEbiography}
[{\includegraphics[width=1in,height=1.3in,clip]{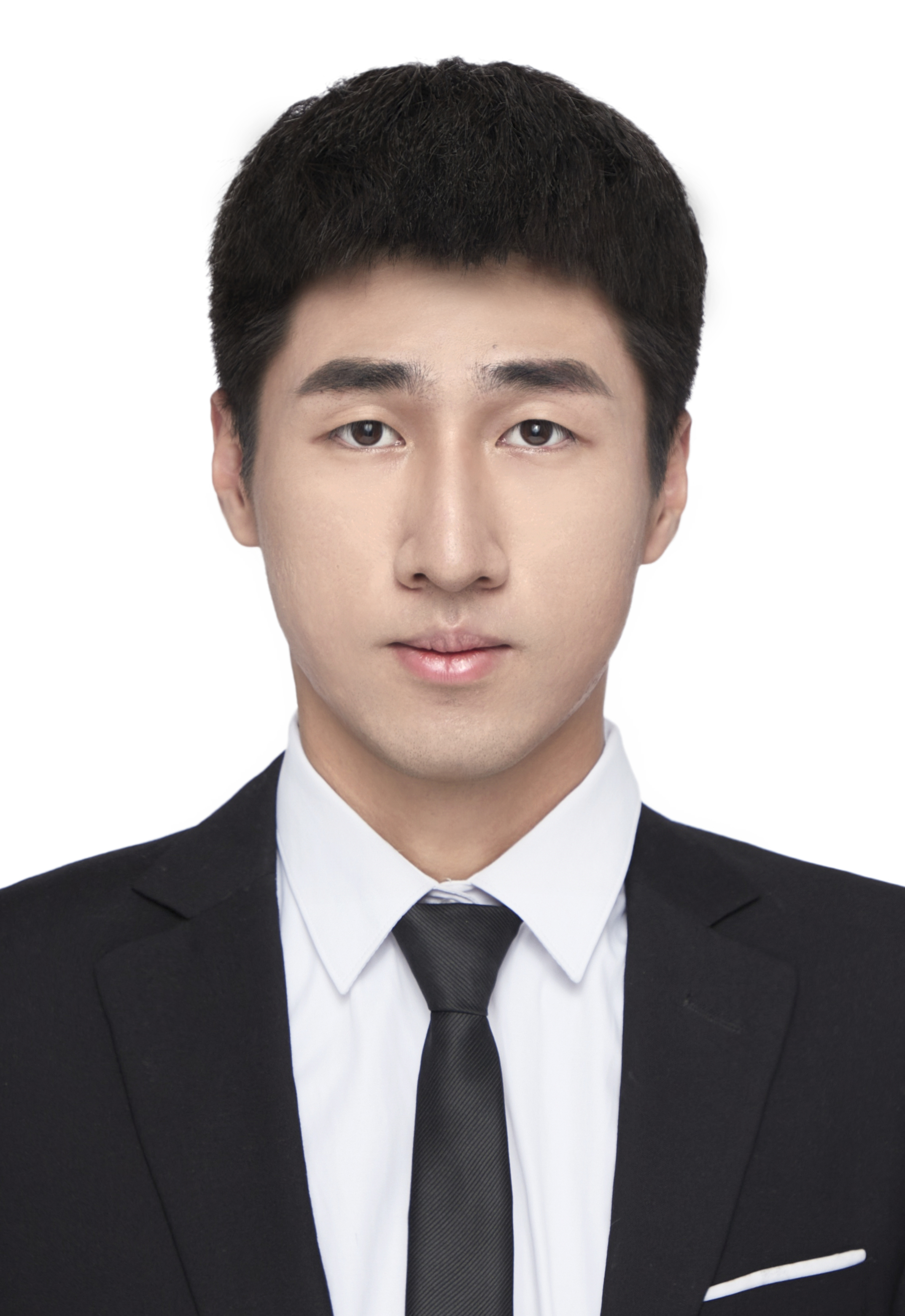}}]{Guangyang Lu} received the B.S.degree in Electronic Information Engineering from the College of Electronic Science and Enginering, Jilin University, Changchun, China, in 2023. He is currently pursuing a Master's degree in the School of Integrated Circuits, Tsinghua University. His research interests  include dataflow architecture, AI accelerator and large-scaling chip design.

\end{IEEEbiography}

\vspace{-40 pt}

\begin{IEEEbiography}
[{\includegraphics[width=1.15in,height=1.35in,clip,keepaspectratio]{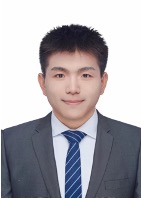}}]{Qize Yang} received the B.S. degree in micro electronic science and technology from Tsinghua University, Beijing, China, in 2022.
He is currently working toward the Ph. D. degree with the School of Integrated Circuits, Tsinghua unversity, Beijing, China. His research interests include distributed learning, wafer scale computing and communication, and neural network acceleration. 

\end{IEEEbiography}

\vspace{-30 pt}

\begin{IEEEbiography}
[{\includegraphics[width=1in,height=1.25in,clip,keepaspectratio]{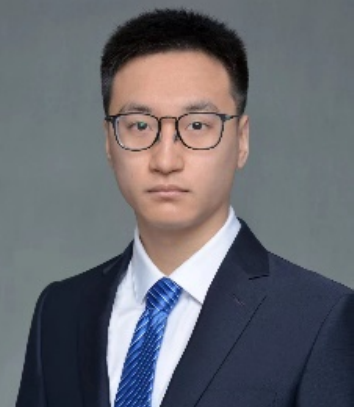}}]{Jiahao Zhang} receiced the B.S. degree in microelectronics science and engineering from the School of Intergrated Circuits, Tsinghua University, Bejing, China, in 2023. He is currently working with Professor Shouyi Yin in the School of Intergrated Circuits, Tsinghua University. His research interests include computer architecture, AI acceleration and processors, and large-scaling chip design.  

\end{IEEEbiography}

\vspace{-30 pt}

\begin{IEEEbiography}
[{\includegraphics[width=1.1in,height=1.3in,clip,keepaspectratio]{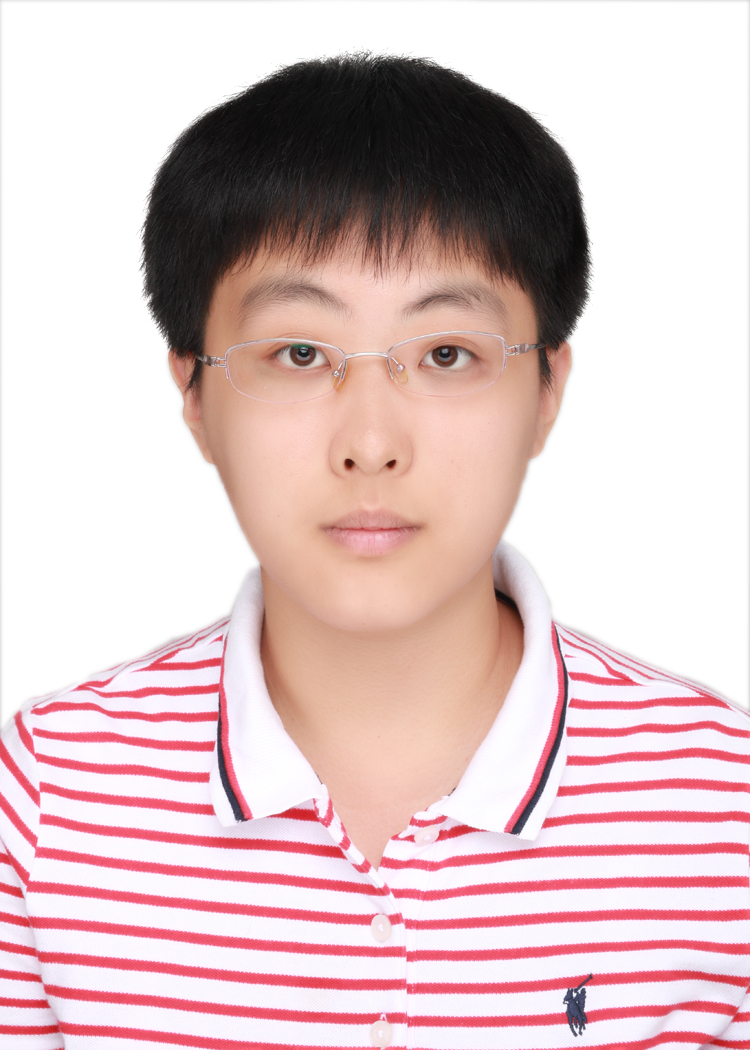}}]{Jinxi Li} received the B.S. degree in electronic engineering from Tsinghua University, Beijing, China in 2023 and is currently working on her Ph.D. degree. Her research interests include software and hardware associated optimization and compiling optimization.  
\end{IEEEbiography}

\vspace{-30 pt}

\begin{IEEEbiography}
[{\includegraphics[width=1in,height=1.25in,clip,keepaspectratio]{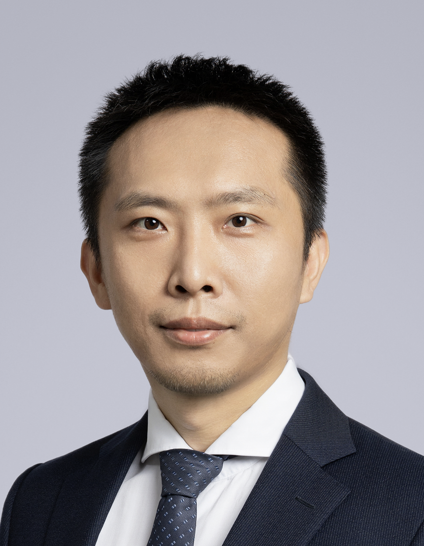}}]{Chao Li} earned the BS degree from Zhejiang University (with honors), China, and the Ph.D. degree from the University of Florida, America. He is currently a Full Professor in the Department of Computer Science and Engineering, Shanghai Jiao Tong University (SJTU). At SJTU, he is a member of Emerging Parallel Computing Center (EPCC) and directs the Sustainable Architecture and Infrastructure Lab (SAIL). 
His main research focus is designing system architectures for energy-efficient, high-performance, extreme-scale computers.

\end{IEEEbiography}

\vspace{-40 pt}

\begin{IEEEbiography}
[{\includegraphics[width=1in,height=1.25in,clip,keepaspectratio]{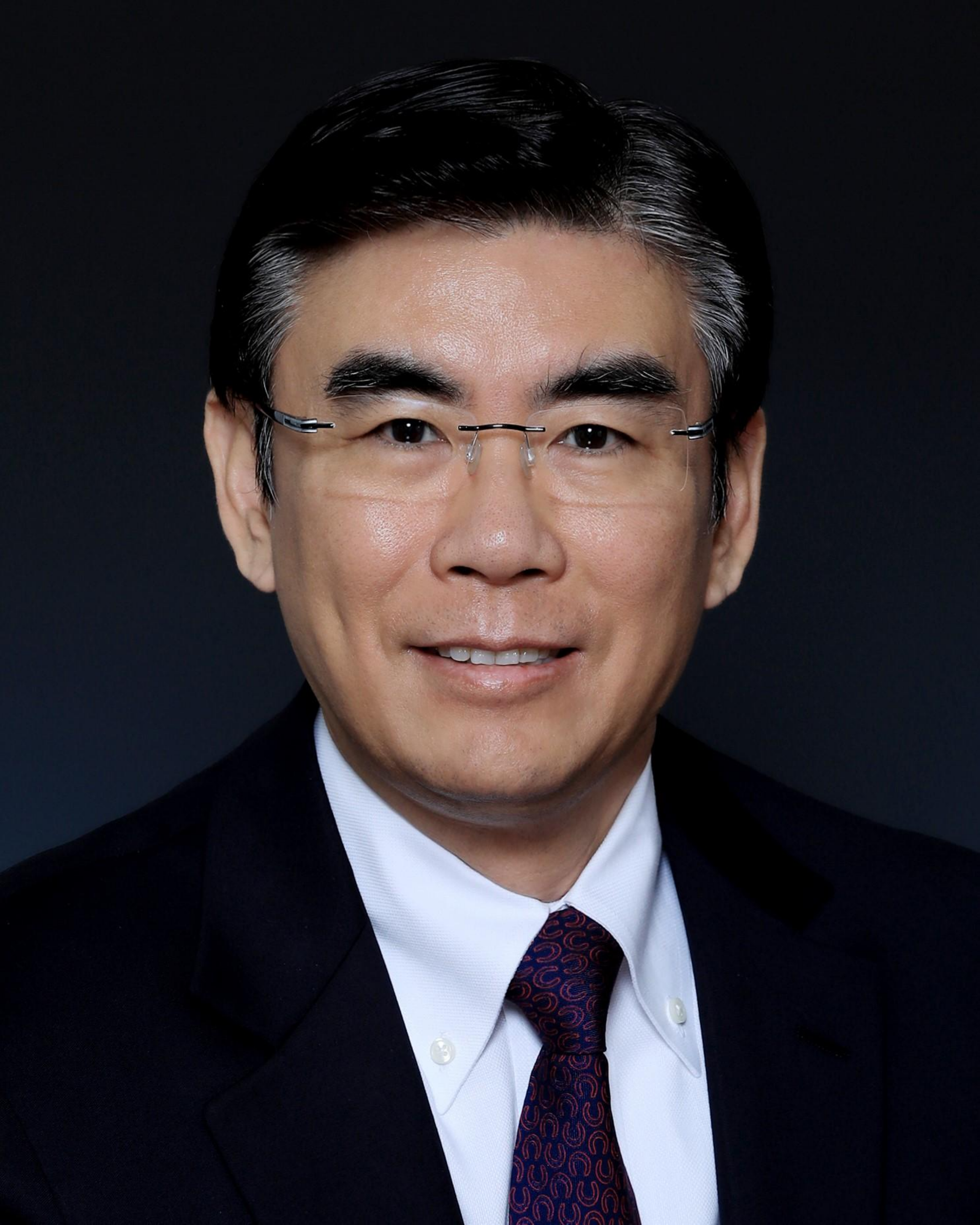}}]{Shaojun Wei} 
(Fellow, IEEE) was born in Beijing, China, in 1958. He received the Ph.D. degree from the Faculte Polytechnique de Mons, Belgium, in 1991. He is currently a Professor with the School of Integrated Circuits, Tsinghua University, Beijing. He is a Senior Member of the Chinese Institute of Electronics (CIE). His main research interests include VLSI SoC design, EDA methodology, and communication ASIC design. 
\end{IEEEbiography}

\vspace{-30 pt}

\begin{IEEEbiography}
[{\includegraphics[width=1in,height=1.25in,clip,keepaspectratio]{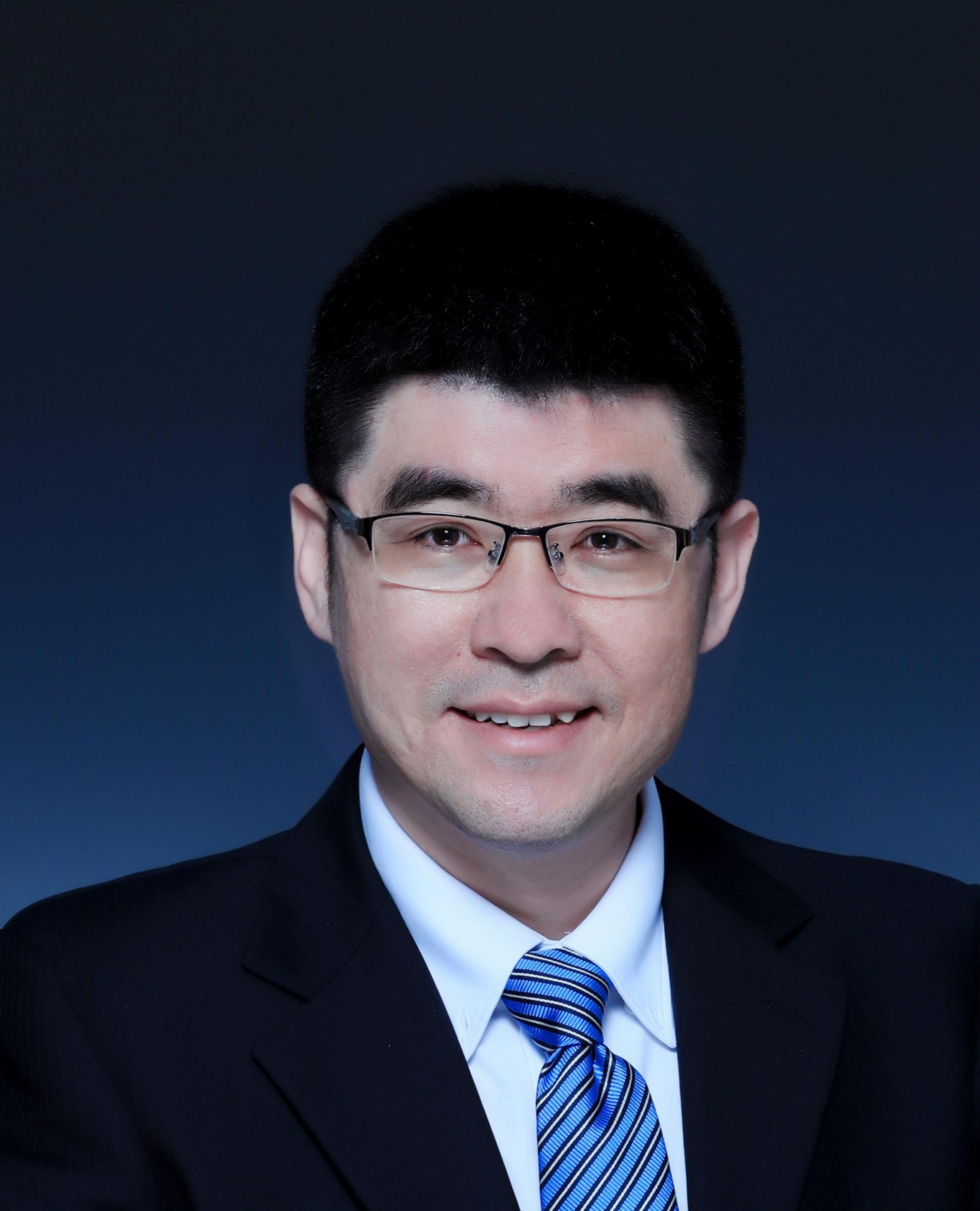}}]{Yang Hu} 
received the B.S. degree from Tianjin University in 2007, the M.S. degree from Tsinghua University in 2011, and the Ph.D. degree from the University of Florida in 2017. 
He was a tenure-track Assistant Professor with the Department of Electrical and Computer Engineering, The University of Texas at Dallas, from 2017 to 2021. 
He is currently an Associate Professor with the School of Integrated Circuits, Tsinghua University. 
His research interests include network-oriented domain-specific architecture, co-optimization of memory/computing system for AI, and hardware security of heterogeneous systems. 
\end{IEEEbiography}

\vspace{-30 pt}

\begin{IEEEbiography}
[{\includegraphics[width=1in,height=1.25in,clip,keepaspectratio]{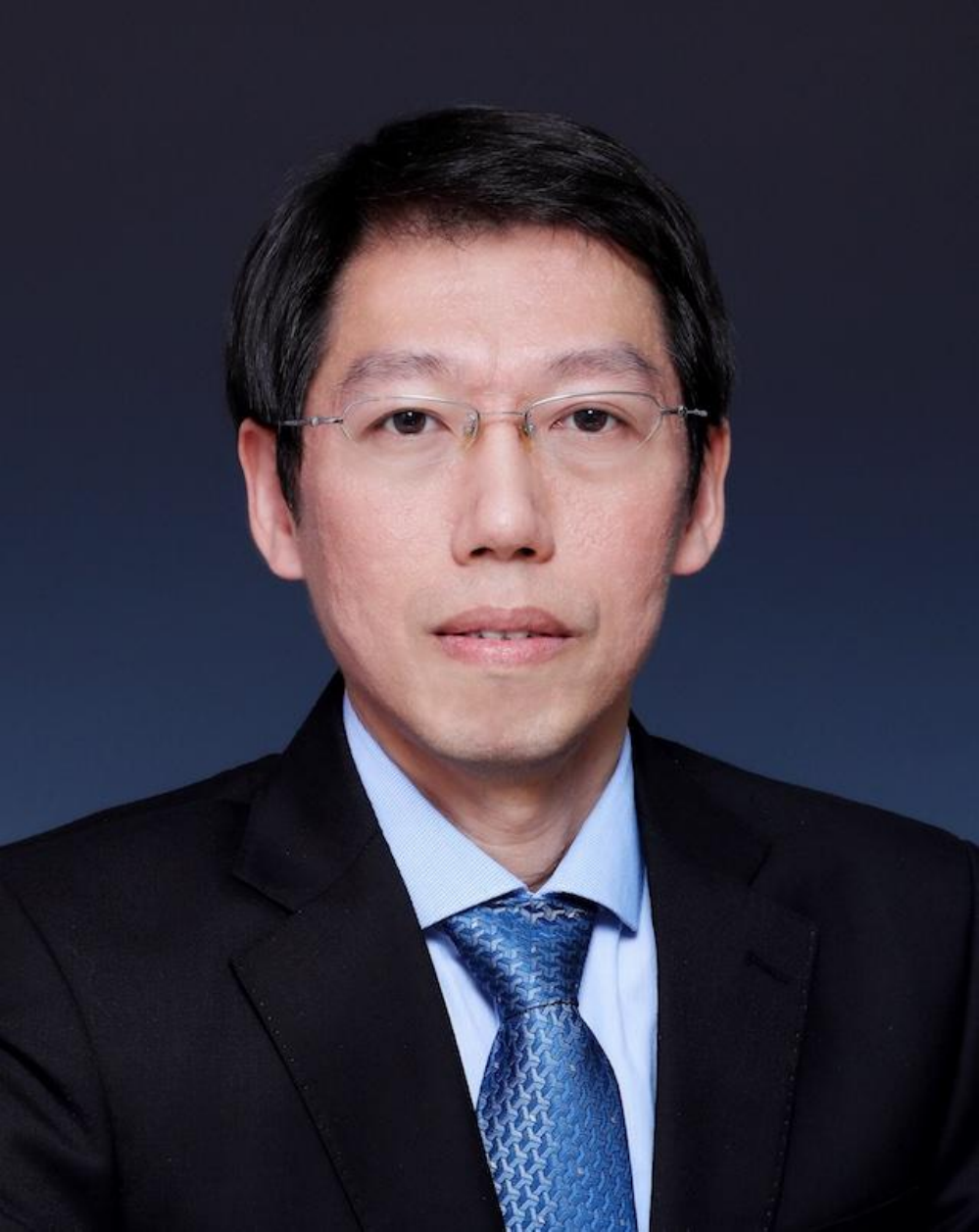}}]{Shouyi Yin} 
(Senior Member, IEEE) received the B.S., M.S., and Ph.D. degrees in electronic engineering from Tsinghua University, Beijing, China, in 2000, 2002, and 2005, respectively. He was a Research Associate with Imperial College London, London, U.K. He is currently a Full Professor and the Vice Director of the School of Integrated Circuits, Tsinghua University. His research interests include reconfigurable computing, AI processors, and high level synthesis. 
\end{IEEEbiography}
\vfill

\end{document}